\documentclass[11pt]{article}

\usepackage[preprint]{acl}

\usepackage{times}
\usepackage{latexsym}
\usepackage{algorithm}
\usepackage{verbatim}  
\usepackage{microtype}
\usepackage{url}
\usepackage{hyperref}
\usepackage{amsmath}
\usepackage{amssymb}
\usepackage{mathtools}
\usepackage{bm}
\usepackage{amsthm}

\usepackage{array}
\usepackage{booktabs}
\usepackage{makecell}
\usepackage{multirow}
\usepackage{colortbl}

\usepackage{graphicx}
\usepackage{xcolor}
\usepackage{tcolorbox}

\usepackage{algorithm}
\usepackage{algorithmicx}
\usepackage{algpseudocode}

\usepackage{pifont}
\usepackage{lineno}
\usepackage{enumitem}
\usepackage{fontawesome}
\usepackage{natbib}
\usepackage{soul}
\usepackage[T1]{fontenc}

\usepackage[utf8]{inputenc}

\usepackage{microtype}

\usepackage{inconsolata}

\usepackage{graphicx}

\title{PAHQ: Accelerating Automated Circuit Discovery through Mixed-Precision Inference Optimization}

\author{
  Xinhai Wang$^{1}$ \quad
  Shu Yang$^{1}$ \quad
  Liangyu Wang$^{1}$ \quad
  Lin Zhang$^{2}$ \quad \\
  Huanyi Xie$^{1}$ \quad
  Lijie Hu$^{1}$ \quad
  Di Wang$^{1}$
  \\\\
  $^{1}$King Abdullah University of Science and Technology \\
  $^{2}$Harbin Institute of Technology \\
  \texttt{\{xinhai.wang, shu.yang, liangyu.wang, huanyi.xie, lijie.hu, di.wang\}@kaust.edu.sa} \\
  \texttt{23s058005@stu.hit.edu.cn} 
}

\begin{document}
\maketitle

\begin{abstract}
Circuit discovery, which involves identifying sparse and task-relevant subnetworks in pre-trained language models, is a cornerstone of mechanistic interpretability. Automated Circuit Discovery (ACDC) has emerged as a pivotal methodology in circuit discovery, but its application to large language models is severely limited by computational inefficiency and prohibitively high memory requirements. Although several accelerated approaches have been proposed, they primarily rely on linear approximations to ACDC, which significantly compromises analytical faithfulness. Our proposed method for accelerating automated circuit discovery, Per Attention Head Quantization (PAHQ), takes a fundamentally different approach by optimizing the efficiency of each individual patching operation. PAHQ leverages a fundamental alignment between activation patching and mixed-precision quantization (MPQ): interpretability analysis through patching essentially performs targeted ablation studies. Therefore, we can maintain high precision exclusively for investigated components while safely reducing precision elsewhere in the network. PAHQ-accelerated ACDC reduces runtime by up to 80\% and memory consumption by up to 30\% compared to unaccelerated ACDC while maintaining faithfulness. 
Importantly, our method readily integrates with existing edge-based circuit discovery techniques by modifying the attention computation mechanism. This training-free approach provides a practical and novel pathway for accelerating mechanistic interpretability methods. Our code is available at \href{https://github.com/626619403/PAHQ}{https://github.com/626619403/PAHQ.}

\end{abstract}

\section{Introduction}

As LLMs scale up~\cite{vaswani2017attention,minaee2025largelanguagemodelssurvey}, understanding their internal mechanisms becomes crucial for effective utilization and supervision~\cite{chen2021evaluating,Li_2022, Manzari_2023}.

A main method for understanding model behavior is to decompose a model into circuits—interpretable computational subgraphs responsible for specific model behaviors~\cite{wang2023interpretability,bereska2024mechanistic}.
The Automatic Circuit Discovery (ACDC) method proposed by~\cite{conmy2023automated} served as the first scalable, end-to-end approach for identifying behavior-related circuits in models. While ACDC is effective, it relies on activation patching (replacing activations from one forward pass with those from another) using an expensive greedy search that ablates each edge, which limits its scalability~\cite{bhaskar2024finding}. One alternative is Edge Attribution Patching (EAP)~\cite{syed2023attribution}, which replaces activation patching with gradient-based linear approximations. While EAP is significantly faster, these first-order approximations often compromise faithfulness to the full model.

Given these computational and memory constraints, along with the need to maintain model performance, quantization~\cite{han2016deep,jacob2017quantization,krishnamoorthi2018quantizing} naturally emerges as a promising approach to address these challenges. The quantization operation is to convert the high-precision floating-point parameters of the model into lower-precision representations to reduce the memory footprint and speed up inference. 
However, in Section~\ref{sec:rtnq}, we found that when precision is reduced from 32 bits to 8 bits, direct quantization (also called Round-to-Nearest Quantization, RTN-Q) significantly increases the minimum quantization step size, causing activation values below this threshold to be truncated to zero and ultimately rendering the automated circuit discovery method ineffective. Furthermore, RTN-Q reduces activation precision when the model aggregates these values (e.g., in residual streams).. When two numbers are added, the larger number dominates the smaller one, masking the activation differences of the selected edge so that the algorithm cannot detect the influence of the selected edge. This represents the most challenging obstacle in applying quantization to accelerate automated circuit discovery methods. 

Fortunately, all these problems stem from the precision degradation of selected edges, and we can use mixed-precision quantization (MPQ)~\cite{saxena2025resqmixedprecisionquantizationlarge, huang2025slimllmsaliencedrivenmixedprecisionquantization, lee2024owqoutlierawareweightquantization} to mitigate these effects. Mixed-precision quantization enables selective maintenance of high numerical precision exclusively for specific components under investigation while safely reducing precision elsewhere in the network, avoiding the issues caused by direct quantization. However,  most conventional mixed-precision quantization methods require substantial pre-training time to identify important components. Moreover, due to GPU memory limitations, mixed-precision quantization methods typically stores high-precision weights in CPU memory, while only the critical weight subset needed for current computation and all low-precision weights are retained on GPU. The dynamic nature of important weights  introduces frequent CPU-GPU weight migration in automated circuit discovery, resulting in significant I/O latency.

Our key insight is that there exists a natural alignment between activation patching~\cite{conmy2023automated} and mixed-precision quantization techniques: activation patching fundamentally conducts targeted ablation by examining how specific neural activations affect model outputs, which perfectly aligns with mixed-precision quantization strategies. Since we already know which components are important when using circuit discovery methods, we can directly leverage the selected edges to eliminate the pre-training process. This characteristic also enables us to predict the next high-precision weights to be accessed, allowing us to effectively mask data transfer overhead through a computation-communication overlap strategy that interleaves low-precision computation with high-precision weight communication, while maintaining optimized memory usage to address the CPU-GPU communication bottleneck.

Motivated by these insights, we develop PAHQ (Per-Attention Head Quantization), a training-free quantization method for automated circuit discovery. We observe that automated circuit discovery algorithms typically evaluate only one edge at a time. We leverage this property by maintaining high precision exclusively for the critical component involved in that edge evaluation, while converting all other components to lower precision. To address the memory scheduling and precision compatibility challenges, we implement a comprehensive CUDA stream scheduler for attention heads that effectively masks high-precision weight loading latency through concurrent low-precision computation.

Across three circuit discovery tasks, ACDC accelerated by PAHQ demonstrated over 40\% improvement in AUC-ROC for classifying circuit component edges compared to 8-bit RTN quantization. When applied to GPT-2, ACDC accelerated by PAHQ reduces computation time by approximately 80\% while achieving memory compression ratios exceeding 30\% compared to unaccelerated ACDC.

\section{Direct Quantization Does Not Work on ACDC}
To better illustrate our proposed approach, we provide necessary preliminaries about automated circuit discovery and quantization in the appendix~\ref{appendix:preliminary}. In this section, we present preliminary studies on directly applying quantization to ACDC to determine whether it can improve ACDC's efficiency. 
\begin{figure}[htb]
    \centering
    \vspace{-10pt}
    \includegraphics[width=0.76\columnwidth]{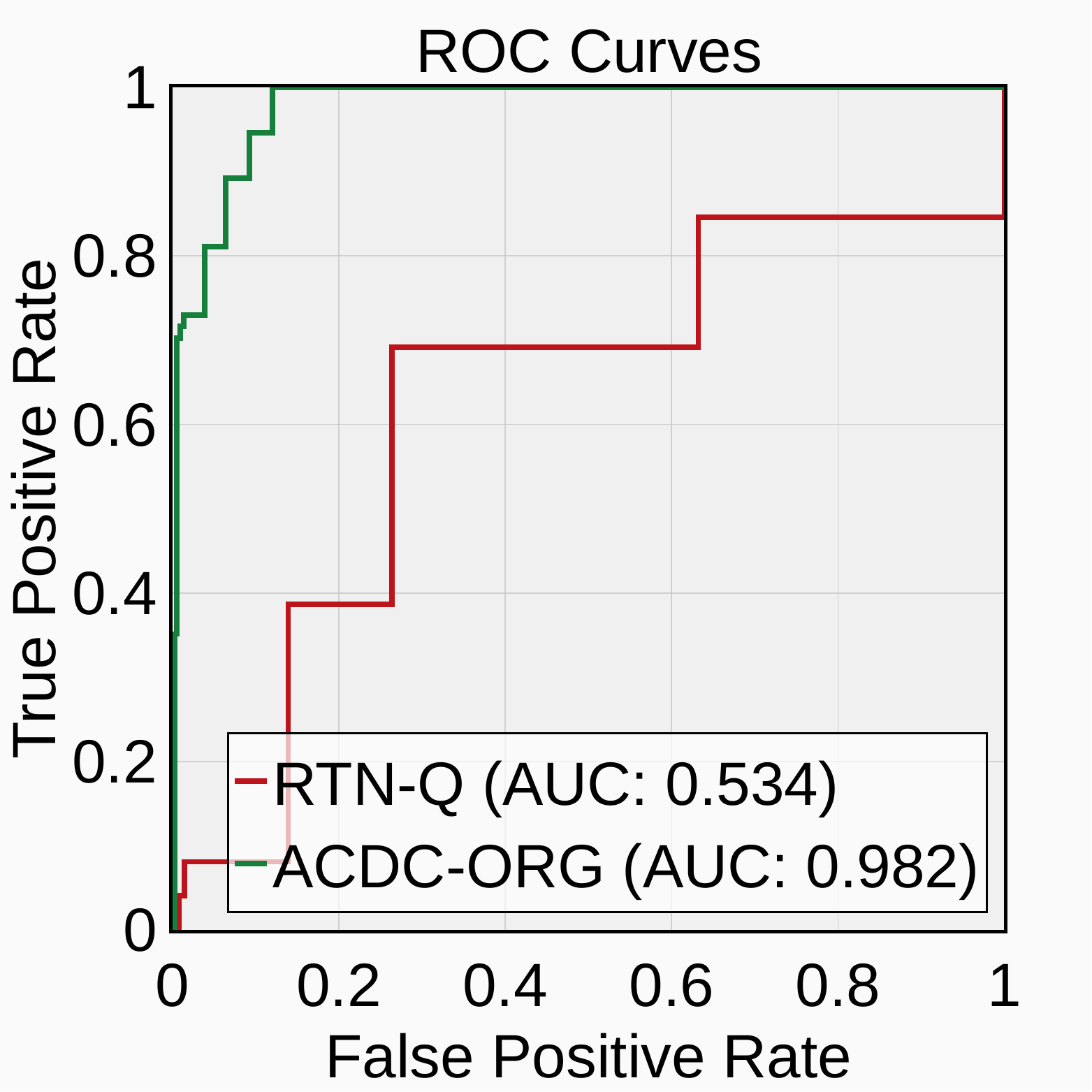}
    \caption{ROC curves. RTN-quantization degrades performance.}
    \label{fig:roc}
\end{figure}

We directly applied quantization from 32 bits to 8 bits to all weights and activations of the model~\cite{radford2019language} in the ACDC~\cite{conmy2023automated} method, which is called Round-to-Nearest Quantization (RTN-Q), on the Indirect Object Identification (IOI) task~\cite{wang2023interpretability} and tested its performance. Figure~\ref{fig:roc} shows that the AUC-ROC was significantly reduced compared to the original ACDC, indicating that direct quantization is not suitable for it. 
As mentioned in Appendix~\ref{circuit discovery}, since the replacement operation only changes the activation value of the selected edge, the difference in the model's output is primarily related to the difference in activation values of the selected edge, rather than other components. Specifically, this degradation can be attributed to two main factors: Numerical Underflow and Mantissa Loss. 
\label{sec:rtnq}

\paragraph{Numerical Underflow.} When the quantization precision of weights and activation values is reduced from 32 bits to 8 bits (or less), the minimum representable precision increases substantially. For FP8\_E4M3\footnote{FP8\_E4M3 is an 8-bit floating point storage format, with 4 bits for the exponent, 3 bits for the mantissa, and one bit for the sign.}, the minimum quantization step size is: $ 2^{-6}$. This means that if the difference between inference outputs before and after replacing activation values of selected edges is less than the minimum quantization step size, this difference will be truncated to zero. For automated circuit discovery methods, the threshold is likely smaller than this value ~\cite{conmy2023automated,syed2023attribution}. Numerical underflow causes the activation value of the selected component or the change in model output to be set to zero when it is smaller than the minimum quantization step size. This causes the automated circuit discovery method to fail in pruning the edge.

\paragraph{Mantissa Loss}
\label{ManDis}
RTN quantization reduces numerical precision when activation values are summed (such as in residual connections). 
When two floating-point numbers are added, the exponents are aligned to the larger number. The mantissa of the smaller number must be right-shifted and discarded until its exponent matches the larger one. Therefore, in the FP8\_E4M3 format, which has only 3 mantissa bits, if the exponent difference is 3 or greater~\cite{kuzmin2024fp8quantizationpowerexponent}, the significant bits of the smaller number are completely lost. It causes the activation difference introduced by the selected edge to be overwhelmed when aggregated with other activation values, preventing the algorithm from detecting the edge's influence on the model output.

\section{Method}

To address the precision degradation and computational bottleneck challenges identified in Section \ref{sec:rtnq}, we introduce PAHQ (Per-Attention Head Quantization), a training-free mixed-precision quantization method specifically designed for automated circuit discovery methods. 

\subsection{Per Attention Head Quantization}

\paragraph{Key Insight: Natural Alignment Between Circuit Discovery and Mixed-Precision} 

As analyzed in Section~\ref{sec:rtnq}, ACDC's edge importance criterion depends primarily on the precision of the selected edge's activation values. ACDC evaluates edge importance through the loss difference $\Delta L(e)$ defined in~\eqref{Loss Function}:{
\small
\begin{equation}
\Delta L(e) = \mathbb{E}_{(\mathbf{z}, \mathbf{z}') \sim \mathcal{T}} \left| L(E_{G \setminus \{e\}}(\mathbf{z}, \mathbf{z}')) - L(E_G(\mathbf{z})) \right|.
\end{equation}}
Where $\mathbf{z}$, $\mathbf{z}'$ are clean prompt and corrupted prompt. 
The critical observation is that this loss difference is dominated by the numerical precision of activations flowing through the edge $e = (u \rightarrow v)$ under evaluation, where $v$ and $u$ are nodes in the computational graph $G$ corresponding to computational components. Specifically, let $\mathbf{a}_u^{(high)}$ and $\mathbf{a}_u^{(low)}$ denote the high-precision and low-precision activations from source node $u$, respectively. The precision-induced error in edge evaluation can be formalized as:
{
\small
\begin{equation}
\epsilon_{precision}(e) = \left| \Delta L(e; \mathbf{a}_u^{(high)}) - \Delta L(e; \mathbf{a}_u^{(low)}) \right|.
\end{equation}
}

Both numerical underflow and mantissa loss issues affect circuit discovery by introducing non-negligible $\epsilon_{precision}(e)$, while other components remain largely unaffected since their activations are not subject to intervention during edge $e$'s evaluation. Our approach leverages a fundamental insight: there exists a natural alignment between activation patching in circuit discovery and mixed-precision quantization strategies. Mixed-precision quantization enables selective maintenance of high numerical precision exclusively for specific components under investigation while safely reducing precision elsewhere in the network, avoiding the issues caused by direct quantization. 

Mixed-precision quantization requires extensive calibration to identify important components, which contradicts the goal of accelerating circuit discovery. However, in automated circuit discovery, we inherently know which components are critical—precisely those edges currently under evaluation. This enables us to eliminate the pre-training phase entirely by directly using the selected edges to guide precision allocation.

\paragraph{Selective Precision Allocation Strategy}

Based on this insight, we propose a dynamic selective precision allocation strategy that adapts to the temporal evaluation sequence. Given the computational graph $G = (V, E)$ from Appendix ~\ref{appendix:preliminary}, we formalize the precision assignment process as follows.

For any edge evaluation at time step $t$, let $e_t = (u_t \rightarrow v_t) \in E$ denote the edge currently under evaluation by ACDC. We define a time-dependent precision allocation function $P_t: V \rightarrow \{\text{FP8}, \text{FP32}\}$ that dynamically assigns precision levels based on the current evaluation context:
{
\small
\begin{equation}
P_t(v) = 
\begin{cases}
\text{FP32}, & \text{if } v = u_t \text{ (source of edge)} \\
\text{FP8}, & \text{otherwise}
\end{cases}
\end{equation}
}
For non-attention components, we uniformly adopt a \texttt{bfloat16} (bf16) quantization setting. This strategy ensures non-critical components operate at reduced precision, thereby preserving the faithfulness of $\Delta L(e_t)$ computation while saving GPU memory.
This strategy ensures that the critical component under investigation maintains full numerical precision while non-critical components operate at reduced precision, thereby preserving the faithfulness of $\Delta L(e_t)$ computation while achieving substantial GPU memory savings.

\paragraph{Hierarchical Weight Scheduling Formulation}

The dynamic precision allocation necessitates sophisticated weight management between CPU and GPU memory hierarchies. Let $\mathbf{W}_{dev(,v)}^{(32)} \in \mathbb{R}^{d \times d}$ and $\mathbf{W}_{dev(,v)}^{(8)} \in \mathbb{Q}^{d \times d}$ denote the high-precision (FP32) and low-precision (FP8) weight matrices for node $v$ on device $dev\in\{CPU, GPU\}$, respectively, where $\mathbb{Q}$ represents the quantized number space.

In conventional mixed-precision quantization, weights are transferred reactively upon demand:
{
\small
\begin{equation}
\mathbf{W}_{GPU}^{(t)} = \mathbf{W}_{CPU}^{(32)} \cdot \mathbb{I}_{P_t(v) = \text{FP32}} + \mathbf{W}_{GPU}^{(8)} \cdot \mathbb{I}_{P_t(v) = \text{FP8}}
\end{equation}
}
where $\mathbb{I}$ denotes the indicator function and the transfer occurs synchronously, creating computational bottlenecks.
However, since ACDC evaluates edges one-by-one in a  sequence $\{e_1, e_2, \ldots, e_{|E|}\}$, we can exploit this structure for {\bf predictive weight scheduling}. Let $S_t = \{v \mid P_t(v) = \text{FP32}\}$ denote the set of nodes requiring high precision at time $t$. Our predictive scheduler implements:
{
\small
\begin{equation}
\text{Prefetch}(t) = \mathbf{W}_{CPU}^{(32)}[S_{t+1}] \xrightarrow{\text{async}} \mathbf{W}_{GPU}^{(32)}[S_{t+1}]
\end{equation}
}
{
\small
\begin{equation}
\text{Compute}(t) = f(\mathbf{x}_t; \mathbf{W}_{GPU}^{(32)}[S_t], \mathbf{W}_{GPU}^{(8)}[V \setminus S_t]),
\end{equation}
}
where $f$ represents the forward pass function and $\mathbf{x}_t$ is the input at time $t$. The prefetch operation for time $t+1$ executes asynchronously during the computation at time $t$, effectively masking the data transfer latency through computation-communication overlap. This predictive scheduling transforms the sequential bottleneck into a parallel pipeline, achieving near-optimal throughput while maintaining the precision guarantees required for accurate circuit discovery.

\subsection{Algorithm Implementation in Practice}

\paragraph{Mixed-Precision Computation Design.} 
Although our theoretical design aims to update only selected attention head weights to 32-bit precision while maintaining the rest at 8-bit precision, GPU hardware constraints impose specific implementation requirements. Due to contiguous memory layout requirements and the constraint that matrix multiplication operands must maintain consistent precision~\cite{micikevicius2018mixedprecisiontraining}, we cannot directly mix weights of different precisions within the same computational kernel.

To address this limitation, we implement a two-phase computation strategy. For a transformer layer $l$ with $H$ attention heads and a target head $h^* \in \{1, 2, \ldots, H\}$, we define the attention computation process as follows.

First, we compute attention scores separately for each precision level. For each attention component $comp \in \{Q, K, V\}$, we generate both low-precision and high-precision outputs:
{
\small
\begin{equation}
\mathbf{A}_{comp,8}^{(l)} = f_{FP8}(\mathbf{T}, \mathbf{W}_{comp,8}^{(l)}) \in \mathbb{Q}^{B \times S \times H \times d_k}
\end{equation}
}
{
\small
\begin{equation}
\mathbf{A}_{comp,32}^{(l,h^*)} = f_{FP32}(\mathbf{T}, \mathbf{W}_{comp,32}^{(l,h^*)}) \in \mathbb{R}^{B \times S \times d_k}
\end{equation}
}
where $B$ denotes batch size, $S=$ sequence length, $d_k=$ key dimension, and $\mathbf{T}$ represents the input tensor. The function $f_{prec}$ with $prec\in \{FP8, FP32\}$ represents attention computation at the specified precision level. 

Next, we perform selective replacement to construct the mixed-precision attention tensor. For each head $h$, we define the final attention scores as:
{
\small
\begin{equation}
\mathbf{A}_{comp}^{(l,h)} = 
\begin{cases}
\mathbf{A}_{comp,32}^{(l,h)}, & \text{if } h = h^* \\
\text{Cast}_{FP32}(\mathbf{A}_{comp,8}^{(l,h)}), & \text{otherwise}
\end{cases}
\label{equ:mixedassembly}
\end{equation}
}
where $\text{Cast}_{FP32}$ converts tensors to FP32 format.
Finally, we unify all activations to FP32 precision to ensure consistency for subsequent matrix operations:
{
\small
\begin{equation}
\mathbf{A}_{comp}^{(l)} = \text{Concat}_{head}([\mathbf{A}_{comp}^{(l,1)}, \mathbf{A}_{comp}^{(l,2)}, \ldots, \mathbf{A}_{comp}^{(l,H)}]).
\end{equation}
}

This approach maintains the numerical precision required for the critical attention head while achieving substantial memory savings through reduced precision for non-critical components.

\paragraph{Scheduler Implementation.} 
To flexibly replace the precision of any attention head, traditional mixed-precision quantization approaches store high-precision versions of all weights in CPU memory and transfer the 32-bit weights of selected attention heads to the GPU as needed. However, for general mixed-precision quantization methods, data transfer between CPU and GPU typically creates a computational bottleneck that can significantly reduce computation speed.

\begin{figure}[htb]
    \centering
    \includegraphics[width=\columnwidth]{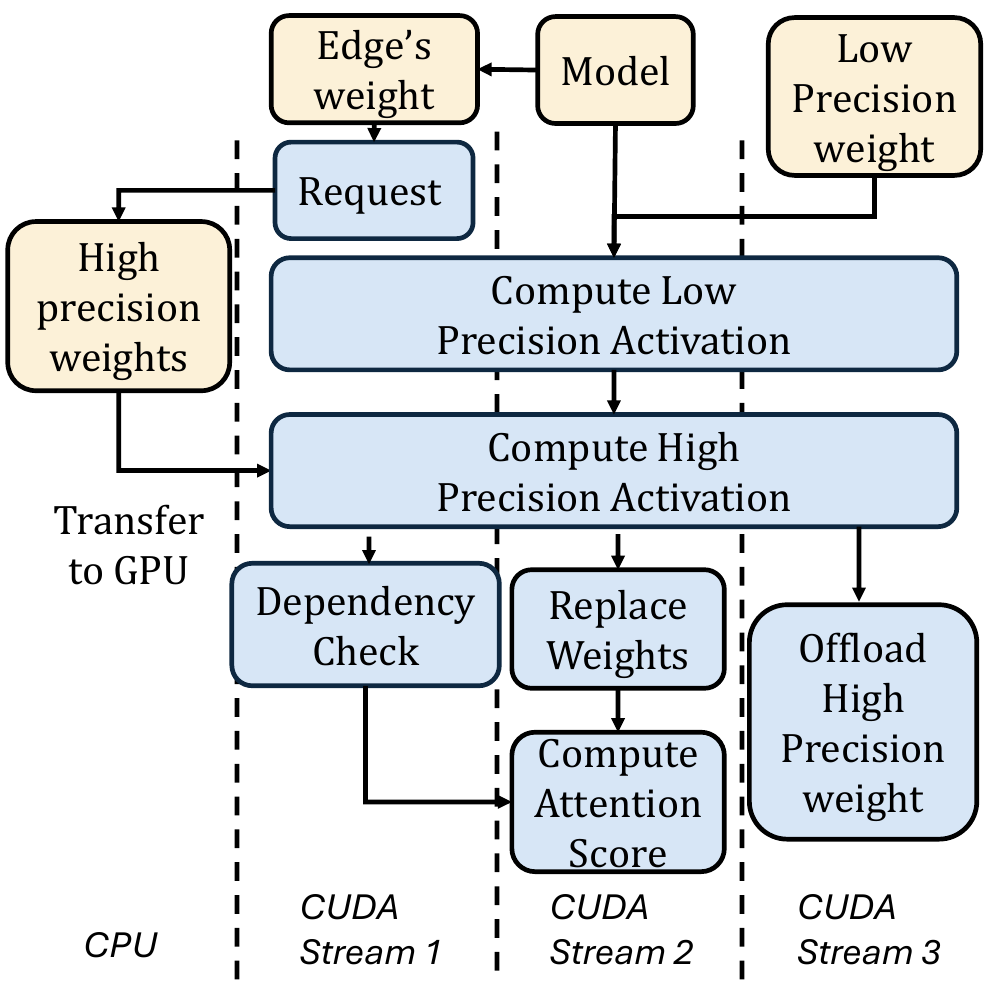}
    \caption{Three-Stream Parallel Scheduler. We overlap high-precision weight transfers with low-precision computations to mask transfer latency.}
    \label{fig:stream}
\end{figure}

To optimize performance, we design a three-stream parallel scheduler that overlaps computation and communication. Let $\mathcal{S} = \{S_{load}, S_{low}, S_{high}\}$ denote the three CUDA streams for weight loading, low-precision computation, and high-precision computation, respectively.

The scheduling process implements the following pipeline for each layer $l$ and target head $h^*$:

\paragraph{Phase 1: Asynchronous Weight Prefetching}
We initiate asynchronous weight transfers from CPU to GPU for both the target attention head and output projection:
{
\small
\begin{equation}
\begin{aligned}
S_{load}: \mathbf{W}_{QKV,32}^{(l,h^*)} &\xrightarrow{async} \text{GPU}, \\
\mathbf{W}_{O,32}^{(l)} &\xrightarrow{async} \text{GPU}
\end{aligned}
\end{equation}
}
where $\mathbf{W}_{QKV,32}^{(l,h^*)},\mathbf{W}_{O,32}^{(l,h^*)}$ represents the combined weight matrices for Query, Key, Value and Output projections of the target attention head $h^*$ in layer $l$ stored in FP32 precision.
\paragraph{Phase 2: Overlapped Computation}

To address the computational bottleneck caused by data transfer between CPU and GPU, we execute low-precision computations for all heads while simultaneously loading and computing high-precision weights for the target head, thereby overlapping computation and memory transfer to minimize idle time. For each attention component $comp \in \{Q, K, V\}$, we execute the following overlapped operations: 
{\small
\begin{align}
S_{low}&: \mathbf{A}_{comp,8}^{(l)} = f_{FP8}(\mathbf{T}, \mathbf{W}_{comp,8}^{(l)}) \quad \text{(async)} \\
&\text{Sync}(S_{load}, \mathbf{W}_{comp,32}^{(l,h^*)}) \\
S_{high}&: \mathbf{A}_{comp,32}^{(l,h^*)} = f_{FP32}(\mathbf{T}, \mathbf{W}_{comp,32}^{(l,h^*)}) \quad \text{(async)}
\end{align}
}
where $Sync(\cdot)$ is a CUDA stream synchronization operation that waits for the specified stream or memory transfer to complete before proceeding to the next operation.
\paragraph{Phase 3: Synchronization and Assembly}
We need a stream to synchronize all parallel streams, combine the mixed-precision attention outputs using the selective replacement strategy, and perform the final attention computation with unified FP32 precision:
{\small
\begin{align}
&\text{Sync}(S_{low}, S_{high}) \\
&\mathbf{A}_{comp}^{(l)} = \text{MixedAssembly}(\mathbf{A}_{comp,8}^{(l)}, \mathbf{A}_{comp,32}^{(l,h^*)}, h^*) \\
&\text{Sync}(S_{load}, \mathbf{W}_{O,32}^{(l)}) \\
&\mathbf{Y} = \text{Attention}(\mathbf{A}_Q^{(l)}, \mathbf{A}_K^{(l)}, \mathbf{A}_V^{(l)}) \cdot \mathbf{W}_{O,32}^{(l)},
\end{align}
}
where the $\text{MixedAssembly}$ function implements the selective replacement defined in Equation~\ref{equ:mixedassembly}.

The weight transfer latency $T_{transfer}$ is masked by low-precision computation time $T_{comp}^{low}$ because the CPU-to-GPU transfer happens in parallel with the low-precision computation. Since both operations occur simultaneously on different streams, the total time is approximately $\max(T_{transfer}, T_{comp}^{low})$ rather than $T_{transfer} + T_{comp}^{low}$, achieving near-optimal throughput. The complete PAHQ-accelerated ACDC algorithm and schematic diagram are provided in Appendix~\ref{appendix:pahq_acdc}.

\section{Experiment}

\subsection{Experimental Setting}
\noindent{\bf Tasks.} We selected three tasks that are commonly studied in automatic circuit discovery: Indirect Object Identification (IOI)~\cite{wang2023interpretability}, Docstring~\cite{heimersheim2023circuit} and Greater Than~\cite{nanda2023progress}. For task introduction, please refer to Appendix~\ref{appendix:detail}.

\noindent {\bf Baselines.} For baselines, we employ ACDC~\cite{conmy2023automated} as our baseline method for quantization and compare the performance of HISP~\cite{michel2019sixteen}, SP~\cite{cao2021low}, EAP~\cite{syed2023attribution}, Edge Pruning~\cite{bhaskar2024finding}, 
RTN quantization on ACDC, original ACDC, and our method. We present the comparison results with Edge Pruning in Appendix~\ref{appendix:edge_pruning}. 

\noindent {\bf Metrics.} Since we apply quantization to the ACDC method in this work, we adopt its original evaluation metric (the Area Under the Receiver Operating Characteristic curve, AUC-ROC) to assess faithfulness. For a more detailed description of metrics, please refer to Appendix~\ref{appendix:detail}. We also assess faithfulness using the metric proposed by ~\cite{hanna2024faith} in Appendix~\ref{appendix:detail}.

\noindent {\bf Models.} The models used are the same as ACDC, including gpt2~\cite{radford2019language}, attn-4l~\cite{heimersheim2023circuit}, and redwood-2l~\cite{Gokaslan2019OpenWeb}. All of these models are downloaded from HuggingFace.

\subsection{Experimental Results}
This section compares the performance of different methods in terms of our primary metrics: faithfulness, task performance, runtime, and memory requirements. 

\begin{table}[htbp]
\centering
\small
\renewcommand{\arraystretch}{1.2}

\begin{tabular}{c*{6}{c}}  
\toprule
\multirow{3}{*}{\textbf{Method}} & \multicolumn{6}{c}{\textbf{AUC-ROC}} \\
\cline{2-7}
 & \multicolumn{2}{c}{\textbf{IOI}} & \multicolumn{2}{c}{\textbf{Greater Than}} & \multicolumn{2}{c}{\textbf{Docstring}} \\
\cline{2-7}
 & \textbf{KL} & \textbf{Task} & \textbf{KL} & \textbf{Task} & \textbf{KL} & \textbf{Task} \\ 
\midrule

ACDC & \textbf{0.98} & \textbf{0.97} & \textbf{0.91} & \underline{0.87} & \textbf{0.95} & \textbf{0.94} \\
RTN-Q & 0.53 & 0.58 & 0.62 & 0.50 & 0.47 & 0.53 \\
HISP & 0.78 & 0.83 & 0.70 & 0.68 & 0.81 & 0.82 \\
SP & 0.82 & 0.72 & 0.80 & 0.81 & 0.83 & 0.81 \\
EAP & 0.71 & 0.70 & 0.83 & \textbf{0.89} & 0.83 & 0.83 \\
\midrule
\textbf{PAHQ} & \underline{0.96} & \underline{0.94} & \underline{0.87} & 0.84 & \underline{0.89} & \underline{0.89} \\ 

\bottomrule
\end{tabular}
\caption{AUC-ROC for different circuit discovery methods on specified tasks and models. The thresholds used were logarithmically evenly distributed between 0.001 and 3.16, for a total of 21 values, consistent with ACDC. ``\textbf{bold}'' and ``\underline{underlined}'' represents the best and second best performance. RTN-Q and PAHQ are abbreviations of RTN-Q-accelerated ACDC and PAHQ-accelerated ACDC.}
\label{tab:auc_roc}
\end{table}

\begin{table*}[t]
\centering
\small

\begin{tabular}{@{}lcc*{9}{c}@{}}
\toprule

\multirow{3}{*}{\textbf{Threshold}} & \multirow{3}{*}{\textbf{Method}} & \multirow{3}{*}{\textbf{Metric}} & \multicolumn{9}{c}{\textbf{Models and Tasks}} \\
\cmidrule{4-12}
 & & & \multicolumn{3}{c}{\textbf{IOI}} & \multicolumn{3}{c}{\textbf{Greater Than}} & \multicolumn{3}{c}{\textbf{Docstring}} \\
\cmidrule(lr){4-6} \cmidrule(lr){7-9} \cmidrule(lr){10-12}
 & & & \textbf{gpt2} & \textbf{4l} & \textbf{2l} & \textbf{gpt2} & \textbf{4l} & \textbf{2l} & \textbf{gpt2} & \textbf{4l} & \textbf{2l} \\
\midrule
\multirow{6}{*}{\textbf{0.001}} & \multirow{2}{*}{\textbf{ACDC}} & \textbf{KL div} & \textbf{0.54} & \textbf{0.45} & \textbf{0.37} & \textbf{0.60} & \textbf{0.53} & \textbf{0.48} & \textbf{0.16} & \textbf{0.06} & \underline{0.03} \\
\cmidrule(lr){3-12}
 & & \textbf{Task} & \textbf{0.54} & \textbf{0.42} & \textbf{0.35} & \textbf{0.64} & \textbf{0.49} & \textbf{0.45} & \textbf{0.17} & \textbf{0.08} & \textbf{0.07} \\
\cmidrule(lr){2-12}
 & \multirow{2}{*}{\makecell{\textbf{RTN-Q}\\\textbf{ACDC}}} & \textbf{KL div} & 0.32 & 0.28 & 0.20 & 0.43 & 0.30 & 0.39 & 0.08 & \underline{0.04} & 0.01 \\
\cmidrule(lr){3-12}
 & & \textbf{Task} & 0.33 & 0.29 & 0.19 & 0.41 & 0.35 & 0.35 & 0.10 & 0.05 & 0.02 \\
\cmidrule(lr){2-12}
 & \multirow{2}{*}{\makecell{\textbf{PAHQ}\\\textbf{ACDC}}} & \textbf{KL div} & \underline{0.51} & \underline{0.44} & \underline{0.35} & \underline{0.58} & \underline{0.48} & \underline{0.46} & \textbf{0.16} & \underline{0.04} & \textbf{0.04} \\
\cmidrule(lr){3-12}
 & & \textbf{Task} & \underline{0.52} & \underline{0.39} & \underline{0.29} & \underline{0.61} & \underline{0.47} & \underline{0.43} & \underline{0.14} & \underline{0.06} & \underline{0.04} \\
\midrule
\multirow{6}{*}{\textbf{0.01}} & \multirow{2}{*}{\textbf{ACDC}} & \textbf{KL div} & \textbf{0.57} & \textbf{0.50} & \textbf{0.41} & \textbf{0.60} & \textbf{0.55} & \underline{0.38} & \underline{0.14} & \textbf{0.09} & \textbf{0.06} \\
\cmidrule(lr){3-12}
 & & \textbf{Task} & \textbf{0.53} & \textbf{0.46} & \underline{0.41} & \textbf{0.65} & \textbf{0.52} & \textbf{0.43} & \textbf{0.15} & \textbf{0.07} & \textbf{0.04} \\
\cmidrule(lr){2-12}
 & \multirow{2}{*}{\makecell{\textbf{RTN-Q}\\\textbf{ACDC}}} & \textbf{KL div} & 0.33 & 0.26 & 0.22 & 0.42 & 0.37 & \textbf{0.41} & 0.11 & \underline{0.08} & 0.03 \\
\cmidrule(lr){3-12}
 & & \textbf{Task} & 0.36 & 0.25 & 0.17 & 0.49 & 0.39 & 0.36 & 0.09 & 0.04 & 0.00 \\
\cmidrule(lr){2-12}
 & \multirow{2}{*}{\makecell{\textbf{PAHQ}\\\textbf{ACDC}}} & \underline{KL div} & \underline{0.56} & \underline{0.49} & \underline{0.39} & \underline{0.58} & \textbf{0.55} & \underline{0.36} & \textbf{0.15} & 0.06 & \underline{0.04} \\
\cmidrule(lr){3-12}
 & & \textbf{Task} & \underline{0.51} & \textbf{0.46} & \textbf{0.43} & \underline{0.60} & \underline{0.48} & \underline{0.39} & \underline{0.12} & \textbf{0.07} & \textbf{0.04} \\
\midrule
\multirow{6}{*}{\textbf{0.1}} & \multirow{2}{*}{\textbf{ACDC}} & \textbf{KL div} & \textbf{0.60} &  \textbf{0.51} &  \underline{0.44} & \textbf{0.65} & \textbf{0.49} & \textbf{0.55} & \textbf{0.14} & \textbf{0.10} & \textbf{0.08} \\
\cmidrule(lr){3-12}
 & & \textbf{Task} & \textbf{0.58} & \textbf{0.53} & \textbf{0.43} & \textbf{0.65} & \textbf{0.54} & \textbf{0.49} & \textbf{0.14} & \textbf{0.06} & \underline{0.03} \\
\cmidrule(lr){2-12}
 & \multirow{2}{*}{\makecell{\textbf{RTN-Q}\\\textbf{ACDC}}} & \textbf{KL div} & 0.32 & 0.29 & 0.24 & 0.50 & 0.46 & 0.41 & 0.08 & 0.05 & 0.03 \\
\cmidrule(lr){3-12}
 & & \textbf{Task} & 0.32 & 0.28 & 0.25 & 0.43 & 0.31 & 0.39 & 0.10 & 0.04 & 0.01 \\
\cmidrule(lr){2-12}
 & \multirow{2}{*}{\makecell{\textbf{PAHQ}\\ \textbf{ACDC}}} & \textbf{KL div} & \underline{0.54} & \underline{0.50} & \textbf{0.45} & \underline{0.63} & \underline{0.48} & \underline{0.45} & \textbf{0.14} & \underline{0.09} & \underline{0.06} \\
\cmidrule(lr){3-12}
 & & \textbf{Task} & \underline{0.56} & \underline{0.48} & \underline{0.40} & \underline{0.57} & \underline{0.51} & \underline{0.42} & \underline{0.12} & \textbf{0.06} & \textbf{0.04} \\
\bottomrule
\end{tabular}
\caption{Accuracies for different quantization methods across various tasks, thresholds, and models. ``Task'' in column ``metric'' means task metric. Attn-4l and Redwood-2l are represented as 2l and 4l. For different metrics and tasks, within each threshold group, the highest value in each column is bolded. When there are no tied highest values, the second highest value is underlined.} 
\label{tab:quantization_compare}
\end{table*}

\paragraph{AUC-ROC Performance Analysis.} As illustrated in Tab.~\ref{tab:auc_roc}, PAHQ-accelerated ACDC outperforms both RTN-Q ACDC and EAP approaches across three tasks, and only has minimal faithfulness loss compared to the original ACDC method. For the IOI task, PAHQ-accelerated ACDC achieves AUC-ROC values of 0.96 and 0.94 when optimized using KL divergence and task metrics, substantially exceeding RTN-Q ACDC and EAP. However, for the Greater Than task, our method falls slightly below the EAP~\cite{syed2023attribution} due to the lower performance ceiling imposed by the original ACDC. And we notice that EAP has bad performance on IOI, which is the same as the results in~\cite{hanna2024faith}.

\paragraph{Accuracy Across Different Thresholds and Models}
Tab.~\ref{tab:quantization_compare} presents a detailed performance comparison of various quantization methods, tasks, thresholds, and model configurations. Our method consistently outperforms RTN-Q while only suffering a small performance loss over unaccelerated ACDC. For example, in the IOI task, when the threshold of the gpt2 is 0.001, PAHQ-accelerated ACDC achieves 0.51 on KL divergence and 0.52 on task metrics, significantly outperforming RTN-Q ACDC while only slightly declining from the original ACDC method. This trend remains consistent across different model sizes (Attn-4l and Redwood-2l). Even on the harder task of Docstring, our method still performs well. We also assess our method's scalability to larger models in Appendix~\ref{appendix:scalability}.

\begin{table}[htbp]
\centering
\small
\renewcommand{\arraystretch}{1.1}
\begin{tabular}{@{}l*{3}{cc}@{}}
\toprule
\multirow{5}{*}{\textbf{Method}} & \multicolumn{6}{c}{\textbf{Performance Comparison}} \\
\cmidrule{2-7}
 & \multicolumn{2}{c}{\textbf{gpt-2}} & \multicolumn{2}{c}{\textbf{attn-4l}} & \multicolumn{2}{c}{\textbf{redwood-2l}} \\
\cmidrule(lr){2-3} \cmidrule(lr){4-5} \cmidrule(l){6-7}
 & \textbf{Time} & \textbf{Mem} & \textbf{Time} & \textbf{Mem} & \textbf{Time} & \textbf{Mem} \\
 & \textbf{(m:s)} & \textbf{(GB)} & \textbf{(m:s)} & \textbf{(GB)} & \textbf{(m:s)} & \textbf{(GB)} \\
\midrule

ACDC & 99:18 & 6.23 & 4:03 & 3.30 & 0:28 & 1.54 \\
RTN-Q & \underline{27:47} & \textbf{4.15} & \underline{1:17} & \textbf{1.96} & \underline{0:09} & \textbf{0.99} \\
\midrule
\textbf{PAHQ} & \textbf{20:36} & \underline{4.24} & \textbf{1:01} & \underline{2.11} & \textbf{0:06} & \underline{1.21} \\

\bottomrule
\end{tabular}

\caption{Runtime and memory comparison of quantization methods on the IOI task (threshold = 0.001). Bold and underlined denote the best and second-best results.}
\vspace{-15pt}
\label{tab:runtime}
\end{table}

\paragraph{Runtime and Memory.} To demonstrate the improvement of the circuit discovery method using PAHQ and its performance relative to the RTN-Q quantization method, we tested the operating results in three cases, as shown in Tab.~\ref{tab:runtime}. Our method performed lower-level stream scheduling and matrix multiplication optimization, reduces about 80\% running time and nearly 1/3 memory usage compared to the original method, close to RTN-Q ACDC. This is because our method only needs to store one additional high-precision attention head compared to the RTN-Q method.

\subsection{Ablation Studies}\label{sec:ablation}

\paragraph{Scheduling.} Tab.~\ref{tab:scheduling} shows adding a scheduler can significantly reducing the running time. When both the weight loading stream and the low/high precision calculation stream are enabled, the running time is only 20 minutes, which is about 79\% less than the worst case. Enabling only the weight loading stream (49 minutes) or the low/high precision calculation stream (72 minutes) can also improve parts of performance. We believe that when only the low/high precision calculation stream is enabled, since the time required for model weight loading is longer than the high precision calculation time, it will frequently wait for weight transfer, so its performance is not as good as when only the weight loading stream is enabled.

\begin{table}[htbp]
\centering
\small

\begin{tabular}{>{\centering\arraybackslash}m{2.5cm}>{\centering\arraybackslash}m{2.5cm}>{\centering\arraybackslash}m{1.5cm}}
\toprule

\makecell{Weight Loading\\Stream} & \makecell{Low/High-Precision\\Computation Stream} & Runtime (m) \\
\midrule
\checkmark & \checkmark &  \textbf{20} \\
\checkmark & \ding{55} & 49 \\
\ding{55} & \checkmark & 72 \\
\ding{55} & \ding{55} & 94 \\
\bottomrule
\end{tabular}

\caption{Impact of scheduler on runtime. Experiments were conducted using the GPT-2 model on the IOI task.}
\vspace{-15pt}
\label{tab:scheduling}
\end{table}

\paragraph{Different Quantization Precisions.} From Tab.~\ref{tab:bits_acc_aucroc}, it is evident that different quantization precisions significantly impact model performance. 4-bit quantization severely compromises model accuracy and faithfulness. We believe that this is because the precise activation value output by the selected edge is passed, the number of bits of the weights and output activation values of subsequent layers is too small, so small numerical changes cannot be retained. Therefore, it is almost impossible to use 4-bit quantization. In contrast, 16-bit quantization offers minimal precision improvements compared to 8-bit. To prove the correctness of our quantization method, we tested the performance of the model under different quantization scales. See Appendix~\ref{appendix:quantization_ablation} for details.

\begin{table}[htbp]
\centering
\small

\begin{tabular}{>{\centering\arraybackslash}m{2cm}>{\centering\arraybackslash}m{2cm}>{\centering\arraybackslash}m{2cm}}
\toprule
Precision & Accuracy & AUC-ROC \\
\midrule
4-bit & 0.06 & 0.61 \\
8-bit & 0.54 & 0.93 \\
16-bit & 0.55 & 0.96 \\
\bottomrule
\end{tabular}
\caption{Impact of different quantization precisions on results. Experiments were conducted with a threshold of 0.001 on the IOI task using the gpt-2 model.}
\label{tab:bits_acc_aucroc}

\end{table}
\section{Related Work}

\paragraph{Circuit Discovery.} Neural networks can be abstracted as computational graphs, with circuits serving as the fundamental computational units and building blocks~\cite{olah2020zoom}. Circuits are subgraphs representing critical components necessary for specific tasks, and circuit discovery tasks utilize task information and network structure~\cite{bereska2024mechanistic} to identify important operations and attribute outputs to specific components~\cite{miller2024transformer}, thereby extracting the model's critical circuits for specific tasks. Existing research has found that this attribution approach can significantly enhance interpretability for designated tasks in small models, such as knowledge-intensive tasks~\cite{meng2022locating}, grammatical functions~\cite{wang2023interpretability}, temporal reasoning tasks ~\cite{hanna2024gpt2}, and others.

However, extending circuit discovery methods to large models remains challenging due to the complexity of causal interventions and the exponential growth of circuit interactions as the number of components increases. Although Automated Circuit Discovery (ACDC) ~\cite{conmy2023automated} automates circuit discovery and ensures high faithfulness through exhaustive recursive activation patching, its quadratic complexity with respect to model components becomes prohibitive for large-scale networks. To address this issue, Edge Attribution Patching (EAP)~\cite{syed2023attribution} achieves linear efficiency via gradient-based edge estimation, but this comes at the cost of approximation errors in multi-layer nonlinear computations. EAP-IG~\cite{hanna2024faith} proposed a new faithfulness metric, and improves edge attribution patching's faithfulness by integrated gradients. Edge Pruning~\cite{bhaskar2024finding} prunes by training the model's edge mask, improving faithfulness at the cost of time. Our method balances the precision-efficiency trade-off by introducing per-attention-head quantization and optimizing the inference process, thereby maintaining high precision while reducing computational costs.

\paragraph{Quantization.}  Quantization is a method of converting high-precision model parameters into low-precision representations, which can significantly reduce model size and accelerate inference ~\cite{gholami2021survey}. However, precision requirements exhibit significant heterogeneity across diverse tasks~\cite{nagel2021white, holavanahali2023analysis},  presenting challenges in determining optimal bit allocations universally. To address this problem, ~\cite{lin2024awq} proposed a quantization framework that selectively preserves critical dimensions; SmoothQuant achieves efficient quantization via channel-wise rescaling operations ~\cite{xiao2023smoothquant}. These approaches require additional hyperparameters that need calibration across datasets and models, imposing substantial computational overhead. To address these limitations, ~\cite{yao2022zeroquant} introduced a zero-shot quantization paradigm without fine-tuning procedures. However, it fails to maintain the interpretability of internal model mechanisms, and circuit discovery tasks fundamentally rely on understanding and preserving these intricate mechanisms.

Inspired by task-specific quantization techniques ~\cite{frantar2023sparsegpt, yuan2023rptq, lee2024owq}, we propose PAHQ, a precision-adaptive approach specifically engineered for circuit discovery methods. PAHQ achieves an optimal balance between performance and compression by maintaining high-precision computation for critical attention heads, delivering superior results without additional training through mechanistic interpretability insights.

\section{Conclusions}
In this research, we introduce PAHQ (Per Attention Head Quantization), an efficient quantization scheme for automated circuit discovery methods. PAHQ leverages the intrinsic characteristics of automated circuit discovery by selectively maintaining high-precision representations of critical attention heads, thereby significantly reducing computational overhead while preserving circuit identification faithfulness. This method requires no additional training and can be used as a plug-and-play solution. Experimental results demonstrate that PAHQ achieves an average reduction of approximately 80\% in time consumption and 30\% in memory usage when implemented within the ACDC framework with only slight performance loss, providing an acceleration approach for mechanistic interpretability research without sacrificing analytical quality.

\newpage
\section*{Limitations}

\paragraph{Applicability to Circuit Discovery Algorithms}
Our proposed method, PAHQ, is designed to accelerate circuit discovery techniques that rely on the precise, sequential evaluation of individual edges within the computational graph. Its effectiveness is most pronounced when applied to algorithms like ACDC, where only one edge's activations are patched per inference pass. This allows for a highly targeted application of high-precision computation, maximizing efficiency gains. However, the scope of PAHQ's applicability does not extend uniformly to all circuit discovery paradigms. For methods such as Edge Attribution Patching (EAP), which approximate ACDC by evaluating the influence of multiple edges within a layer simultaneously during a single backward pass, the benefits of our approach would be diminished. To maintain faithfulness, such methods would require holding all concurrently evaluated edges and their associated components at high precision, which would significantly reduce the promised improvements in memory consumption and runtime.

\paragraph{Scalability to Large-Scale Models}
This work demonstrates significant performance improvements on models commonly used in mechanistic interpretability research, such as the GPT-2 family. It is important to acknowledge that a shared limitation across most current circuit discovery methods is their primary application to these relatively small-scale models. While PAHQ serves as an effective accelerator, it primarily mitigates the exponential growth in computational cost associated with the underlying discovery algorithms as model size increases. It does not, however, fundamentally solve the scalability challenge. The core problem of applying exhaustive, intervention-based analyses to foundation models with tens or hundreds of billions of parameters remains an open and critical research area. Our method makes the application to moderately larger models more feasible but does not yet provide a complete solution for today's largest models.

\paragraph{Reliance on Empirical Validation}
The justification for our mixed-precision strategy, particularly regarding its ability to preserve the fidelity of patched activations through subsequent computations, is primarily supported by strong empirical results. We demonstrate through extensive experiments that maintaining high precision for a single critical component is sufficient for the ACDC algorithm to function correctly. However, the paper provides limited theoretical analysis on how the high-precision signal avoids degradation as it propagates through downstream layers operating at lower precision. Furthermore, the choice to configure non-attention components with \texttt{bfloat16} (bf16) precision was determined empirically through experimental testing, rather than being derived from a formal theoretical analysis. A more profound analytical investigation into the distinct precision requirements of different component types (e.g., attention vs. MLP layers) could yield further insights and potentially lead to more refined optimization strategies.
\bibliography{custom}

\begin{thebibliography}{45}
\providecommand{\natexlab}[1]{#1}

\bibitem[{Bereska and Gavves(2024)}]{bereska2024mechanistic}
Lukas Bereska and Efstratios Gavves. 2024.
\newblock \href {https://arxiv.org/abs/2404.14082} {Mechanistic interpretability for ai safety--a review}.
\newblock \emph{arXiv preprint arXiv:2404.14082}.

\bibitem[{Bhaskar et~al.(2024)Bhaskar, Wettig, Friedman, and Chen}]{bhaskar2024finding}
Adithya Bhaskar, Alexander Wettig, Dan Friedman, and Danqi Chen. 2024.
\newblock Finding transformer circuits with edge pruning.
\newblock \emph{Advances in Neural Information Processing Systems}, 37:18506--18534.

\bibitem[{Cao et~al.(2021)Cao, Sanh, and Rush}]{cao2021low}
Steven Cao, Victor Sanh, and Alexander~M Rush. 2021.
\newblock Low-complexity probing via finding subnetworks.
\newblock \emph{arXiv preprint arXiv:2104.03514}.

\bibitem[{Chen et~al.(2021)Chen, Tworek, Jun, Yuan, de~Oliveira~Pinto, Kaplan, Edwards, Burda, Joseph, Brockman, Ray, Puri, Krueger, Petrov, Khlaaf, Sastry, Mishkin, Chan, Gray, Ryder, Pavlov, Power, Kaiser, Bavarian, Winter, Tillet, Such, Cummings, Plappert, Chantzis, Barnes, Herbert-Voss, Guss, Nichol, Paino, Tezak, Tang, Babuschkin, Balaji, Jain, Saunders, Hesse, Carr, Leike, Achiam, Misra, Morikawa, Radford, Knight, Brundage, Murati, Mayer, Welinder, McGrew, Amodei, McCandlish, Sutskever, and Zaremba}]{chen2021evaluating}
Mark Chen, Jerry Tworek, Heewoo Jun, Qiming Yuan, Henrique~Ponde de~Oliveira~Pinto, Jared Kaplan, Harri Edwards, Yuri Burda, Nicholas Joseph, Greg Brockman, Alex Ray, Raul Puri, Gretchen Krueger, Michael Petrov, Heidy Khlaaf, Girish Sastry, Pamela Mishkin, Brooke Chan, Scott Gray, and 39 others. 2021.
\newblock \href {https://arxiv.org/abs/2107.03374} {Evaluating large language models trained on code}.
\newblock \emph{Preprint}, arXiv:2107.03374.

\bibitem[{Conmy et~al.(2023)Conmy, Mavor-Parker, Lynch, Heimersheim, and Garriga-Alonso}]{conmy2023automated}
Arthur Conmy, Augustine~N. Mavor-Parker, Aengus Lynch, Stefan Heimersheim, and Adrià Garriga-Alonso. 2023.
\newblock \href {https://arxiv.org/abs/2304.14997} {Towards automated circuit discovery for mechanistic interpretability}.
\newblock \emph{arXiv preprint arXiv:2304.14997}.

\bibitem[{Fawcett(2006)}]{fawcett2006introduction}
Tom Fawcett. 2006.
\newblock \href {https://doi.org/10.1016/j.patrec.2005.10.010} {Introduction to roc analysis}.
\newblock \emph{Pattern Recognition Letters}, 27:861--874.

\bibitem[{Finlayson et~al.(2021)Finlayson, Mueller, Gehrmann et~al.}]{finlayson2021causal}
Mark Finlayson, Aaron Mueller, Sebastian Gehrmann, and 1 others. 2021.
\newblock \href {https://arxiv.org/abs/2106.06087} {Causal analysis of syntactic agreement mechanisms in neural language models}.
\newblock \emph{arXiv preprint arXiv:2106.06087}.

\bibitem[{Frantar and Alistarh(2023)}]{frantar2023sparsegpt}
Elias Frantar and Dan Alistarh. 2023.
\newblock \href {https://arxiv.org/abs/2301.00774} {Sparsegpt: Massive language models can be accurately pruned in one-shot}.
\newblock \emph{Preprint}, arXiv:2301.00774.

\bibitem[{Gholami et~al.(2021)Gholami, Kim, Dong, Yao, Mahoney, and Keutzer}]{gholami2021survey}
Amir Gholami, Sehoon Kim, Zhen Dong, Zhewei Yao, Michael~W. Mahoney, and Kurt Keutzer. 2021.
\newblock \href {https://arxiv.org/abs/2103.13630} {A survey of quantization methods for efficient neural network inference}.
\newblock \emph{arXiv preprint arXiv:2103.13630}.

\bibitem[{Gokaslan and Cohen(2019)}]{Gokaslan2019OpenWeb}
Aaron Gokaslan and Vanya Cohen. 2019.
\newblock Openwebtext corpus.
\newblock \url{http://Skylion007.github.io/OpenWebTextCorpus}.

\bibitem[{Han et~al.(2016)Han, Mao, and Dally}]{han2016deep}
Song Han, Huizi Mao, and William~J. Dally. 2016.
\newblock \href {https://arxiv.org/abs/1510.00149} {Deep compression: Compressing deep neural networks with pruning, trained quantization and huffman coding}.
\newblock \emph{Preprint}, arXiv:1510.00149.

\bibitem[{Hanna et~al.(2024{\natexlab{a}})Hanna, Liu, and Variengien}]{hanna2024gpt2}
M.~Hanna, O.~Liu, and A.~Variengien. 2024{\natexlab{a}}.
\newblock How does gpt-2 compute greater-than?: Interpreting mathematical abilities in a pre-trained language model.
\newblock In \emph{Advances in Neural Information Processing Systems}, volume~36. NeurIPS.

\bibitem[{Hanna et~al.(2024{\natexlab{b}})Hanna, Pezzelle, and Belinkov}]{hanna2024faith}
Michael Hanna, Sandro Pezzelle, and Yonatan Belinkov. 2024{\natexlab{b}}.
\newblock \href {https://arxiv.org/abs/2403.17806} {Have faith in faithfulness: Going beyond circuit overlap when finding model mechanisms}.
\newblock \emph{Preprint}, arXiv:2403.17806.

\bibitem[{Heimersheim and Janiak(2023)}]{heimersheim2023circuit}
Stefan Heimersheim and Jett Janiak. 2023.
\newblock \href {https://www.alignmentforum.org/posts/u6KXXmKFbXfWzoAXn/a-circuit-for-python-docstrings-in-a-4-layer-attention-only} {A circuit for python docstrings in a 4-layer attention-only transformer}.
\newblock \emph{Alignment Forum}.
\newblock Online; accessed March 2025.

\bibitem[{Holavanahali(2023)}]{holavanahali2023analysis}
Anusha Holavanahali. 2023.
\newblock \emph{An Analysis of Heterogeneous Quantization Schemes for Neural Networks}.
\newblock Rochester Institute of Technology.

\bibitem[{Huang et~al.(2025)Huang, Qin, Liu, Li, Liu, Liu, Benini, Magno, Zhang, and Qi}]{huang2025slimllmsaliencedrivenmixedprecisionquantization}
Wei Huang, Haotong Qin, Yangdong Liu, Yawei Li, Qinshuo Liu, Xianglong Liu, Luca Benini, Michele Magno, Shiming Zhang, and Xiaojuan Qi. 2025.
\newblock \href {https://arxiv.org/abs/2405.14917} {Slim-llm: Salience-driven mixed-precision quantization for large language models}.
\newblock \emph{Preprint}, arXiv:2405.14917.

\bibitem[{Jacob et~al.(2017)Jacob, Kligys, Chen, Zhu, Tang, Howard, Adam, and Kalenichenko}]{jacob2017quantization}
Benoit Jacob, Skirmantas Kligys, Bo~Chen, Menglong Zhu, Matthew Tang, Andrew Howard, Hartwig Adam, and Dmitry Kalenichenko. 2017.
\newblock \href {https://arxiv.org/abs/1712.05877} {Quantization and training of neural networks for efficient integer-arithmetic-only inference}.
\newblock \emph{arXiv preprint arXiv:1712.05877}.

\bibitem[{Krishnamoorthi(2018)}]{krishnamoorthi2018quantizing}
Raghuraman Krishnamoorthi. 2018.
\newblock \href {https://arxiv.org/abs/1806.08342} {Quantizing deep convolutional networks for efficient inference: A whitepaper}.
\newblock \emph{Preprint}, arXiv:1806.08342.

\bibitem[{Kullback and Leibler(1951)}]{kullback1951information}
Solomon Kullback and Richard~A Leibler. 1951.
\newblock On information and sufficiency.
\newblock \emph{The annals of mathematical statistics}, 22(1):79--86.

\bibitem[{Kuzmin et~al.(2024)Kuzmin, Baalen, Ren, Nagel, Peters, and Blankevoort}]{kuzmin2024fp8quantizationpowerexponent}
Andrey Kuzmin, Mart~Van Baalen, Yuwei Ren, Markus Nagel, Jorn Peters, and Tijmen Blankevoort. 2024.
\newblock \href {https://arxiv.org/abs/2208.09225} {Fp8 quantization: The power of the exponent}.
\newblock \emph{Preprint}, arXiv:2208.09225.

\bibitem[{Lee et~al.(2024{\natexlab{a}})Lee, Jin, Kim, Kim, and Park}]{lee2024owqoutlierawareweightquantization}
Changhun Lee, Jungyu Jin, Taesu Kim, Hyungjun Kim, and Eunhyeok Park. 2024{\natexlab{a}}.
\newblock \href {https://arxiv.org/abs/2306.02272} {Owq: Outlier-aware weight quantization for efficient fine-tuning and inference of large language models}.
\newblock \emph{Preprint}, arXiv:2306.02272.

\bibitem[{Lee et~al.(2024{\natexlab{b}})Lee, Jin, Kim, Kim, and Park}]{lee2024owq}
Changhun Lee, Jungyu Jin, Taesu Kim, Hyungjun Kim, and Eunhyeok Park. 2024{\natexlab{b}}.
\newblock \href {https://arxiv.org/abs/2306.02272} {Owq: Outlier-aware weight quantization for efficient fine-tuning and inference of large language models}.
\newblock \emph{Preprint}, arXiv:2306.02272.

\bibitem[{Li et~al.(2022)Li, Choi, Chung, Kushman, Schrittwieser, Leblond, Eccles, Keeling, Gimeno, Dal~Lago, Hubert, Choy, de~Masson~d’Autume, Babuschkin, Chen, Huang, Welbl, Gowal, Cherepanov, Molloy, Mankowitz, Sutherland~Robson, Kohli, de~Freitas, Kavukcuoglu, and Vinyals}]{Li_2022}
Yujia Li, David Choi, Junyoung Chung, Nate Kushman, Julian Schrittwieser, Rémi Leblond, Tom Eccles, James Keeling, Felix Gimeno, Agustin Dal~Lago, Thomas Hubert, Peter Choy, Cyprien de~Masson~d’Autume, Igor Babuschkin, Xinyun Chen, Po-Sen Huang, Johannes Welbl, Sven Gowal, Alexey Cherepanov, and 7 others. 2022.
\newblock \href {https://doi.org/10.1126/science.abq1158} {Competition-level code generation with alphacode}.
\newblock \emph{Science}, 378(6624):1092–1097.

\bibitem[{Lin et~al.(2024)Lin, Tang, Tang, Yang, Chen, Wang, Xiao, Dang, Gan, and Han}]{lin2024awq}
Ji~Lin, Jiaming Tang, Haotian Tang, Shang Yang, Wei-Ming Chen, Wei-Chen Wang, Guangxuan Xiao, Xingyu Dang, Chuang Gan, and Song Han. 2024.
\newblock \href {https://arxiv.org/abs/2306.00978} {Awq: Activation-aware weight quantization for llm compression and acceleration}.
\newblock \emph{Preprint}, arXiv:2306.00978.

\bibitem[{Manzari et~al.(2023)Manzari, Ahmadabadi, Kashiani, Shokouhi, and Ayatollahi}]{Manzari_2023}
Omid~Nejati Manzari, Hamid Ahmadabadi, Hossein Kashiani, Shahriar~B. Shokouhi, and Ahmad Ayatollahi. 2023.
\newblock \href {https://doi.org/10.1016/j.compbiomed.2023.106791} {Medvit: A robust vision transformer for generalized medical image classification}.
\newblock \emph{Computers in Biology and Medicine}, 157:106791.

\bibitem[{Meng et~al.(2022)Meng, Bau, Andonian, Belinkov, Saparov, Torralba, and Cox}]{meng2022locating}
Kevin Meng, David Bau, Alex Andonian, Yonatan Belinkov, Aslan Saparov, Antonio Torralba, and David Cox. 2022.
\newblock Locating and editing factual associations in gpt.
\newblock \emph{Advances in Neural Information Processing Systems}, 35:17359--17372.

\bibitem[{Michel et~al.(2019)Michel, Levy, and Neubig}]{michel2019sixteen}
Paul Michel, Omer Levy, and Graham Neubig. 2019.
\newblock Are sixteen heads really better than one?
\newblock \emph{Advances in neural information processing systems}, 32.

\bibitem[{Micikevicius et~al.(2018)Micikevicius, Narang, Alben, Diamos, Elsen, Garcia, Ginsburg, Houston, Kuchaiev, Venkatesh, and Wu}]{micikevicius2018mixedprecisiontraining}
Paulius Micikevicius, Sharan Narang, Jonah Alben, Gregory Diamos, Erich Elsen, David Garcia, Boris Ginsburg, Michael Houston, Oleksii Kuchaiev, Ganesh Venkatesh, and Hao Wu. 2018.
\newblock \href {https://arxiv.org/abs/1710.03740} {Mixed precision training}.
\newblock \emph{Preprint}, arXiv:1710.03740.

\bibitem[{Miller et~al.(2024)Miller, Chughtai, and Saunders}]{miller2024transformer}
Joseph Miller, Bilal Chughtai, and William Saunders. 2024.
\newblock \href {https://arxiv.org/abs/2407.08734} {Transformer circuit faithfulness metrics are not robust}.
\newblock \emph{Preprint}, arXiv:2407.08734.

\bibitem[{Minaee et~al.(2025)Minaee, Mikolov, Nikzad, Chenaghlu, Socher, Amatriain, and Gao}]{minaee2025largelanguagemodelssurvey}
Shervin Minaee, Tomas Mikolov, Narjes Nikzad, Meysam Chenaghlu, Richard Socher, Xavier Amatriain, and Jianfeng Gao. 2025.
\newblock \href {https://arxiv.org/abs/2402.06196} {Large language models: A survey}.
\newblock \emph{Preprint}, arXiv:2402.06196.

\bibitem[{Nagel et~al.(2021)Nagel, Fournarakis, Amjad, Bondarenko, van Baalen, and Blankevoort}]{nagel2021white}
Markus Nagel, Marios Fournarakis, Rana~Ali Amjad, Yelysei Bondarenko, Mart van Baalen, and Tijmen Blankevoort. 2021.
\newblock \href {https://arxiv.org/abs/2106.08295} {A white paper on neural network quantization}.
\newblock \emph{Preprint}, arXiv:2106.08295.

\bibitem[{Nagel et~al.(2019)Nagel, van Baalen, Blankevoort, and Welling}]{nagel2019data}
Markus Nagel, Mart van Baalen, Tijmen Blankevoort, and Max Welling. 2019.
\newblock \href {https://arxiv.org/abs/1906.04721} {Data-free quantization through weight equalization and bias correction}.
\newblock \emph{arXiv preprint arXiv:1906.04721}.

\bibitem[{Nanda and Bloom(2022)}]{nanda2022transformerlens}
Neel Nanda and Joseph Bloom. 2022.
\newblock Transformerlens.
\newblock \url{https://github.com/TransformerLensOrg/TransformerLens}.

\bibitem[{Nanda et~al.(2023)Nanda, Chan, Lieberum, Smith, and Steinhardt}]{nanda2023progress}
Neel Nanda, Lawrence Chan, Tom Lieberum, Jess Smith, and Jacob Steinhardt. 2023.
\newblock \href {https://arxiv.org/abs/2301.05217} {Progress measures for grokking via mechanistic interpretability}.
\newblock \emph{arXiv preprint arXiv:2301.05217}.

\bibitem[{Olah et~al.(2020)Olah, Cammarata, Schubert, Goh, Petrov, and Carter}]{olah2020zoom}
Chris Olah, Nick Cammarata, Ludwig Schubert, Gabriel Goh, Michael Petrov, and Shan Carter. 2020.
\newblock \href {https://doi.org/10.23915/distill.00024.001} {Zoom in: An introduction to circuits}.
\newblock \emph{Distill}.
\newblock Https://distill.pub/2020/circuits/zoom-in.

\bibitem[{Radford et~al.(2019)Radford, Wu, Child, Luan, Amodei, Sutskever et~al.}]{radford2019language}
Alec Radford, Jeffrey Wu, Rewon Child, David Luan, Dario Amodei, Ilya Sutskever, and 1 others. 2019.
\newblock Language models are unsupervised multitask learners.
\newblock \emph{OpenAI blog}, 1(8):9.

\bibitem[{Saxena et~al.(2025)Saxena, Sharify, Roy, and Wang}]{saxena2025resqmixedprecisionquantizationlarge}
Utkarsh Saxena, Sayeh Sharify, Kaushik Roy, and Xin Wang. 2025.
\newblock \href {https://arxiv.org/abs/2412.14363} {Resq: Mixed-precision quantization of large language models with low-rank residuals}.
\newblock \emph{Preprint}, arXiv:2412.14363.

\bibitem[{Syed et~al.(2023)Syed, Rager, and Conmy}]{syed2023attribution}
Aaquib Syed, Can Rager, and Arthur Conmy. 2023.
\newblock \href {https://openreview.net/forum?id=tiLbFR4bJW} {Attribution patching outperforms automated circuit discovery}.
\newblock In \emph{NeurIPS Workshop on Attributing Model Behavior at Scale}.

\bibitem[{Vaswani et~al.(2017)Vaswani, Shazeer, Parmar, Uszkoreit, Jones, Gomez, Kaiser, and Polosukhin}]{vaswani2017attention}
Ashish Vaswani, Noam Shazeer, Niki Parmar, Jakob Uszkoreit, Llion Jones, Aidan~N. Gomez, Lukasz Kaiser, and Illia Polosukhin. 2017.
\newblock \href {https://arxiv.org/abs/1706.03762} {Attention is all you need}.
\newblock \emph{arXiv preprint arXiv:1706.03762}.

\bibitem[{Vig et~al.(2020)Vig, Gehrmann, Belinkov, Qian, Marchisio, Coogan, Nevo, Shieber, Hewitt, Drozd, and Sanh}]{vig2020investigating}
Jesse Vig, Sebastian Gehrmann, Yonatan Belinkov, Sharon Qian, Kelly Marchisio, Allison Coogan, Daniel Nevo, Stuart~M. Shieber, John Hewitt, Aleksandr Drozd, and Victor Sanh. 2020.
\newblock Investigating gender bias in language models using causal mediation analysis.
\newblock In \emph{Advances in Neural Information Processing Systems}, volume~33, pages 12388--12401.

\bibitem[{Wang et~al.(2022)Wang, Variengien, Conmy, Shlegeris, and Steinhardt}]{wang2023interpretability}
Kevin Wang, Alexandre Variengien, Arthur Conmy, Buck Shlegeris, and Jacob Steinhardt. 2022.
\newblock \href {https://arxiv.org/abs/2211.00593} {Interpretability in the wild: a circuit for indirect object identification in gpt-2 small}.
\newblock \emph{Preprint}, arXiv:2211.00593.

\bibitem[{Xiao et~al.(2023)Xiao, Lin, Seznec, Wu, Demouth, and Han}]{xiao2023smoothquant}
Guangxuan Xiao, Ji~Lin, Mickael Seznec, Hao Wu, Julien Demouth, and Song Han. 2023.
\newblock Smoothquant: Accurate and efficient post-training quantization for large language models.
\newblock In \emph{International Conference on Machine Learning}, pages 38087--38099. PMLR.

\bibitem[{Yao et~al.(2022)Yao, Yazdani~Aminabadi, Zhang, Wu, Li, and He}]{yao2022zeroquant}
Zhewei Yao, Reza Yazdani~Aminabadi, Minjia Zhang, Xiaoxia Wu, Conglong Li, and Yuxiong He. 2022.
\newblock Zeroquant: Efficient and affordable post-training quantization for large-scale transformers.
\newblock \emph{Advances in Neural Information Processing Systems}, 35:27168--27183.

\bibitem[{Yuan et~al.(2023)Yuan, Niu, Liu, Liu, Wang, Shang, Sun, Wu, Wu, and Wu}]{yuan2023rptq}
Zhihang Yuan, Lin Niu, Jiawei Liu, Wenyu Liu, Xinggang Wang, Yuzhang Shang, Guangyu Sun, Qiang Wu, Jiaxiang Wu, and Bingzhe Wu. 2023.
\newblock Rptq: Reorder-based post-training quantization for large language models.
\newblock \emph{arXiv preprint arXiv:2304.01089}.

\bibitem[{Zhang et~al.(2025)Zhang, Dong, Zhang et~al.}]{zhang2025eapgp}
L.~Zhang, W.~Dong, Z.~Zhang, and 1 others. 2025.
\newblock {EAP-GP}: Mitigating saturation effect in gradient-based automated circuit identification.
\newblock \emph{arXiv preprint}, arXiv:2502.06852.

\end{thebibliography}
\appendix

\section{Preliminaries}
\label{appendix:preliminary}
\paragraph{Circuit Discovery.}

A transformer model operates as a computational graph~\cite{olah2020zoom} $G$, where nodes $v$ correspond to computational components (attention heads and MLP layers), and edges $e$ define activation pathways between components. $V$ and $E$ are the collections of all nodes $v$ and edges $e$ in the graph, respectively. Each node $v$ receives inputs as the sum of outputs from its predecessors $\{ u \mid (u \to v) \in E \}$, propagating from input embeddings to the final unembedding layer.

A circuit $C \subseteq G$ is defined as a minimal subgraph that preserves task-critical computation paths from inputs to logits ~\cite{olah2020zoom}. To identify such circuits, \textit{activation patching}~\cite{vig2020investigating, finlayson2021causal} is used to intervene in the computational graph and analyze the contribution of different components. Specifically, given a task distribution $\mathcal{T}$ and input pairs $(\mathbf{z}, \mathbf{z}')$ (clean prompt $\mathbf{z}$ vs. corrupted $\mathbf{z}'$), we construct hybrid activations $E_C(\mathbf{z}, \mathbf{z}')$ by replacing non-circuit edges in $G(\mathbf{z})$ with their counterparts from $G(\mathbf{z}')$. The optimal circuit $C^*$ minimizes the expected loss discrepancy~\cite{zhang2025eapgp}:

\begin{equation}
\label{Circuit Function}
C^* = \arg \min_{C \subseteq G} \mathbb{E}_{(\mathbf{z}, \mathbf{z}') \sim \mathcal{T}} \left| L(E_C(\mathbf{z}, \mathbf{z}')) - L(E_G(\mathbf{z})) \right|,
\end{equation}
where $L(\mathbf{A})$ measures task-specific loss over logits. To automate this process, ACDC~\cite{conmy2023automated} prunes edges with insufficient causal impact:

\begin{equation}
C = G \setminus \{e \in G \mid \Delta L(e) < \tau\},
\end{equation}

\begin{equation}\label{Loss Function}
\Delta L(e) = \mathbb{E}_{(\mathbf{z}, \mathbf{z}') \sim \mathcal{T}} \left| L(E_{G \setminus \{e\}}(\mathbf{z}, \mathbf{z}')) - L(E_G(\mathbf{z})) \right|,
\end{equation}

where $\tau$ is a pruning threshold. While ACDC guarantees precision, its $O(|E|)$ complexity becomes prohibitive for large models.

EAP~\cite{syed2023attribution} circumvents this via gradient-based approximation: for an edge $e = (u \to v)$ with clean/corrupted activations $\mathbf{x}_u, \mathbf{x}'_u$, it estimates:
\begin{equation}
\label{EAP}
L(\mathbf{x}_u) - L(\mathbf{x}'_u) \approx (\mathbf{x}_u - \mathbf{x}'_u) \cdot \frac{\partial L(\mathbf{x}'_u)}{\partial \mathbf{x}_v},
\end{equation}
Though EAP reduces cost to $O(1)$, its reliance on first-order approximations introduces errors when nonlinear interactions dominate ~\cite{wang2023interpretability}.

\label{circuit discovery}

\paragraph{Quantization.}

Quantization  ~\cite{jacob2017quantization,nagel2019data,gholami2021survey} reduces the bit-width of weights and activations to improve computational efficiency, \textbf{trading off some precision for speed.} Consider a group of weights $\mathbf{w}$, the linear operation is $\mathbf{y} = \mathbf{w}\mathbf{x}$, and its corresponding operation after quantization is $\mathbf{y} = Q(\mathbf{w})\mathbf{x}$. Formally, RTN (Round-to-Nearest), the most commonly used quantization method, can be formulated as follows:
\begin{equation}
\label{RTN-Q}
Q(\mathbf{w}) = \Delta \cdot \text{Round}(\frac{\mathbf{w}}{\Delta}), \quad \Delta = \frac{\max(|\mathbf{w}|)}{2^{N-1}}
\end{equation}
Where $N$ is the number of quantization bits and $\Delta$ is the quantization scaling factor determined by the absolute maximum value.

\section{Experiment Details}

\label{appendix:detail}
\subsection{Environment}
All our experiments were conducted on a server equipped with NVIDIA H20-NVLink GPU (96GB) and 16 vCPU AMD EPYC 9K84 96-Core Processor. For low-precision operations, we utilize the FP8\_E4M3 format.

\subsection{Tasks}
\label{tasks}
We selected 3 tasks from the ACDC paper: Indirect Object Identification (IOI), Docstring, and Greater Than:

\begin{itemize}
\item \textbf{IOI}~\cite{wang2023interpretability}: This task tests a model's ability to identify the indirect object in sentences like "When Mary and John went to the store, John gave a gift to Mary." The model must correctly identify "Mary" as the recipient of the gift. The dataset consists of 1,000 examples with varying names and contexts, balanced across different sentence structures.

\item \textbf{Docstring}~\cite{heimersheim2023circuit}: This task evaluates a model's ability to generate appropriate Python function documentation from function signatures and implementations. The dataset comprises 10,000 Python functions from diverse libraries, with varying complexity and domain specificity.

\item \textbf{Greater Than}~\cite{nanda2023progress}: This task evaluates a model's numerical reasoning by asking it to determine which of two numbers is larger. The dataset includes 2,000 pairs of numbers with varying digit lengths and magnitudes, testing both simple and complex numerical comparisons.

\end{itemize}

Since the original ACDC paper provides datasets in non-natural language token form, which are not interoperable between models with different tokenizers, we first convert all data to natural language form and then transform them using the respective model's tokenizer. 

\subsection{Metrics}
\label{Metrics}
For faithfulness metrics across all three tasks, we follow the approach in the ACDC paper by scanning across a range of ACDC thresholds $\tau$, SP regularization parameters $\lambda$, or the number of pruned HISP elements $k$. We plot pessimistic line segments between points on the Pareto frontier of TPR and FPR, covering this range of thresholds~\cite{fawcett2006introduction}.

We employ the Area Under the Receiver Operating Characteristic curve(AUC-ROC) to evaluate faithfulness, as high true positive rates (TPR) and low false positive rates (FPR) conceptually correspond to the method's ability to identify subgraphs corresponding to the underlying algorithm implemented by the neural network and the method's ability to avoid including components that do not participate in eliciting the behavior, respectively. Following the original paper, we also evaluate the performance of these methods in terms of task accuracy, faithfulness, runtime, and memory footprint when using either KL divergence~\cite{kullback1951information} or task-specific metrics as the loss function.

\paragraph{Faithfulness Metric}
\label{appendix:FaithfulnessMetric}

Michael Hanna et al.~\cite{hanna2024faith} pointed out that in the moderate overlap range (approximately $0.5$--$0.6$), different circuits exhibit widely varying faithfulness, and overlap alone cannot predict model behavior; even in single-task settings, circuits whose node overlap with the human benchmark reaches $90\%$ can have faithfulness as low as $0\%$ or only $51\%$, indicating that high overlap does not necessarily correspond to high faithfulness.

They attribute this phenomenon to the fact that overlap metrics ignore critical components: even if the overall overlap is high, omission of important nodes or edges can cause faithfulness to collapse; moreover, overlap metrics are inherently symmetric and cannot capture the asymmetric nature of faithfulness. Even adopting asymmetric overlap measures such as recall still fails to reliably predict actual faithfulness. Consequently, in addition to using the metrics from the original paper, we also employ the faithfulness evaluation metric proposed in~\cite{hanna2024faith} to comprehensively assess our quantized method. The result is shown in Tab~\ref{tab:faithfullness_metric}:

\begin{table}[htbp]
\centering

\begin{tabular}{lccc}

\toprule
  & \multicolumn{3}{c}{\textbf{Task}} \\
\midrule
\textbf{Method} & \textbf{IOI} & \textbf{Docstring} & \textbf{Greater-Than}  \\
\midrule
ACDC &  0.92 & 0.90 & 0.95 \\
RTN-Q & 0.27 & 0.31 & 0.40 \\
PAHQ & 0.85 & 0.84 & 0.91  \\
\bottomrule
\end{tabular}
\caption{The faithfulness result. To value the faithfulness fairly, we used the metric proposed by~\cite{hanna2024faith}. All of experiments run with threshold=0.01. RTN-Q and PAHQ are the abbreviations of RTN-Q-accelerated ACDC and PAHQ-accelerated ACDC.}
\label{tab:faithfullness_metric}
\end{table}

As can be seen from the table, after quantization by our method, the performance of ACDC only decreases slightly, while the results after quantization by RTN-Q decrease seriously.

\subsection{Task Metrics}
\label{TaskMetrics}
For detailed definitions of task metrics and comparative studies with other metrics (such as KL divergence, performance matching, and methods based on performance changes), please refer to Appendices C and E.2 of~\cite{conmy2023automated}. These sections thoroughly discuss the advantages and disadvantages of various metrics, and why KL divergence might be superior to logit difference in most of cases.

In this research, we employed logit difference as the sole task metric for our optimization process. This metric calculates the difference in logit values between the correct answer and distractors in specific tasks. 
\subsection{Edge Cutting Process}

\begin{figure*}[ht]
    \centering
    \includegraphics[width=0.9\textwidth]{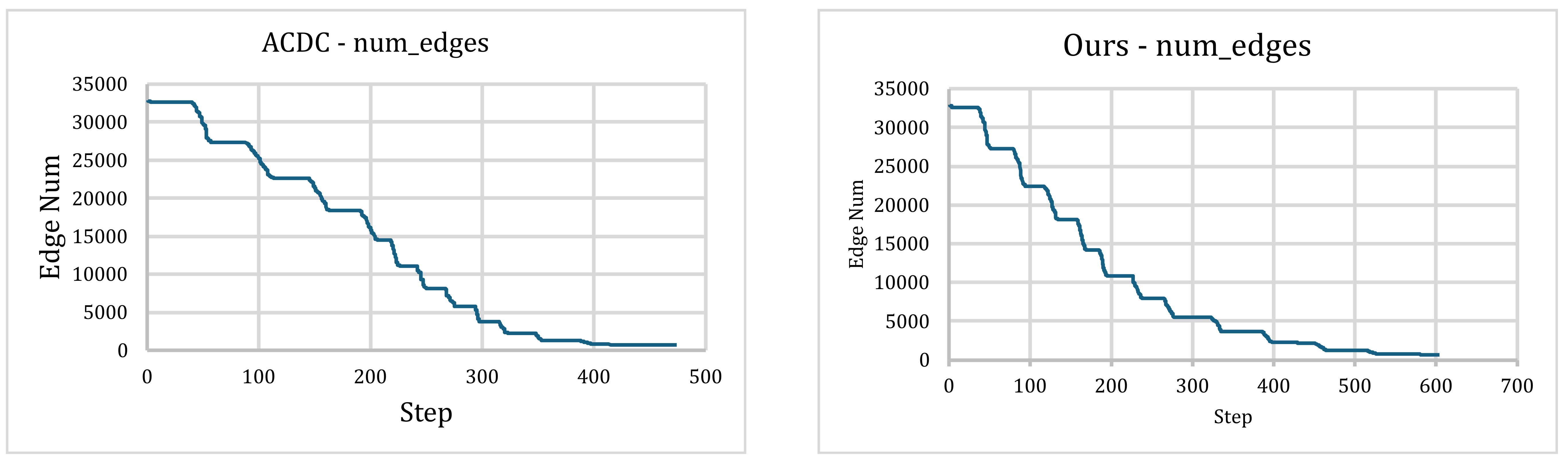}
    \caption{A line graph showing the change in the number of edges in the ACDC circuit before and after quantization using our method as a function of step size. The graph demonstrates that after applying our quantization method, the ACDC circuit does not experience significant changes in terms of sparsity or pruning speed. Although the number of algorithm steps slightly increases, our method still far exceeds the original algorithm in terms of speed.}
    \label{fig:edgenum}
\end{figure*}

As shown in Figure~\ref{fig:edgenum}, the left graph displays the line chart of circuit edge count changes with step size in the original ACDC algorithm, while the right graph shows the results after applying our quantization method. From the comparison, it is clear that both methods exhibit very similar trends during the circuit pruning process. Both graphs start with approximately 35,000 edges, and as the number of steps increases, the edge count decreases in a similar pattern. Notably, although our method increases the total number of steps to about 700 (compared to 500 steps in the original method), the overall sparsification pattern remains consistent, indicating that the quantization process does not significantly alter the sparsity characteristics or pruning efficiency of ACDC. This demonstrates that our quantization method maintains the core performance of the original algorithm while only slightly affecting the computational process, and as previously mentioned, still far exceeds the original algorithm in terms of speed. This minor increase in step count is an acceptable trade-off, as it results in significant performance improvements.

\section{Scalability to Larger Models}
\label{appendix:scalability}

Our main experimental results focus on GPT-2-small, which is the standard benchmark used in the majority of circuit discovery literature~\cite{hanna2024faith, syed2023attribution, bhaskar2024finding} for computational efficiency and fair comparison with existing methods. To demonstrate PAHQ's scalability, we conducted additional experiments on larger GPT-2 variants.

\subsection{Experimental Setup}

We evaluated PAHQ on GPT-2-medium, GPT-2-large, and GPT-2-XL using the IOI task with a threshold of 0.01. Due to computational constraints on larger models, we implemented a batched version of edge evaluation. All experiments were conducted on NVIDIA H20. For reference, ACDC achieves a KL divergence of 0.79 on the IOI task with GPT-2-small.

\subsection{Results}

Tab.~\ref{tab:larger_models} presents the performance of PAHQ across different model scales, compared with the Edge Attribution Patching (EAP) baseline~\cite{syed2023attribution}.

\begin{table}[htbp]
\centering

\begin{tabular}{lccc}
\toprule
\textbf{Model} & \makecell{\textbf{Batch}\\ \textbf{Size}}  & \makecell{\textbf{KL Div}\\ \textbf{(PAHQ)}} & \makecell{\textbf{KL Div}\\ \textbf{(EAP)}} \\
\midrule
GPT-2-medium & 6  & 0.85 & 3.20 \\
GPT-2-large  & 5      & 0.93 & 3.64 \\
GPT-2-XL     & 4       & 0.88 & 3.92 \\
\bottomrule

\end{tabular}
\caption{Performance comparison on larger GPT-2 variants. Lower KL divergence indicates better performance.}
\label{tab:larger_models}
\end{table}

The results demonstrate that PAHQ scales effectively to larger transformer models while maintaining competitive performance. Across all model scales, PAHQ significantly outperforms the EAP baseline. Notably, on all of these models, PAHQ achieves a average KL divergence of 0.89, which represents only a modest degradation compared to ACDC's performance on GPT-2-small (0.79), while providing substantial computational speedup.

The original ACDC method becomes computationally prohibitive on larger models, even with batch processing optimizations; therefore, we are unable to provide ACDC results on larger models.

\section{Comparison with Edge Pruning}
\label{appendix:edge_pruning}

This appendix provides a detailed comparison between PAHQ and Edge Pruning~\cite{bhaskar2024finding}, addressing the relationship between training steps, dataset size, and performance in circuit discovery methods.

\subsection{Methodological Considerations}

An important consideration regarding Edge Pruning's implementation concerns the relationship between training steps and dataset size. The original Edge Pruning paper reports that runtime "barely increases" as dataset size grows. However, this occurs because the method uses a fixed number of training steps (3,000) regardless of dataset size, as confirmed in their official implementation\footnote{See \url{https://github.com/princeton-nlp/Edge-Pruning/blob/main/src/prune/fpt2_ioi.py} for details.}

This design choice means that Edge Pruning processes exactly 3,000 mini-batches regardless of whether the training set contains 100, 10,000, or 100,000,000 examples. Consequently, the method does not reflect how Edge Pruning would behave if allowed to run until convergence on larger datasets. Due to this implementation detail and concerns about fair comparison, we did not include Edge Pruning as a primary baseline in our main experimental results. Instead, we provide this comprehensive analysis to address the relationship between training budget and performance.

\subsection{Experimental Setup}

We evaluated Edge Pruning under multiple regimes of training-step budgets and dataset sizes to ensure a comprehensive comparison. All experiments were conducted on the same hardware (a single NVIDIA H20 GPU) using the IOI task. Hyperparameters for Edge Pruning matched those in the original paper, with the learning rate maintained at 0.8 to avoid instability issues observed at higher rates.

\subsection{Results}

Table~\ref{tab:edge_pruning_comparison} presents the performance comparison across different dataset sizes and training steps. Each cell shows KL divergence and runtime in the format "KL \& Runtime".

\begin{table}[htbp]
\centering
\footnotesize
\renewcommand{\arraystretch}{1.2}

\begin{tabular}{@{}ll*{4}{c}@{}}

\toprule
\multirow{3}{*}{\makecell{\textbf{Dataset}\\ \textbf{Size}}} & \multirow{3}{*}{\textbf{Metric}} & \multicolumn{4}{c}{\textbf{Training Steps}} \\
\cmidrule{3-6}
 & & \textbf{400} & \textbf{800} & \textbf{1600} & \textbf{3000} \\
 & & \textbf{Steps} & \textbf{Steps} & \textbf{Steps} & \textbf{Steps} \\
\midrule

\multirow{2}{*}{200} & KL div & 3.56 & 2.38 & 0.46 & 0.23 \\
 & Time & 31 min & 57 min & 112min & 199min \\
\cmidrule{3-6}

\multirow{2}{*}{400} & KL div & 3.83 & 1.72 & 0.48 & 0.27 \\
 & Time & 29 min & 63min & 98min & 208min \\
\cmidrule{3-6}

\multirow{2}{*}{100,000} & KL div & 4.17 & 2.41 & 0.39 & 0.19 \\
 & Time & 35 min & 52 min & 133min & 212min \\

\midrule
\multirow{2}{*}{\makecell{\textbf{PAHQ} \\ \textbf{ACDC}}} & KL div & \multicolumn{4}{c}{\textbf{0.83}} \\
 & Time & \multicolumn{4}{c}{\textbf{22 min}} \\

\bottomrule
\end{tabular}

\caption{Performance comparison between Edge Pruning and PAHQ across different dataset sizes and training steps.}
\label{tab:edge_pruning_comparison}
\end{table}

A key practical consideration is computational resource usage. PAHQ completes circuit discovery in 22 minutes using only 5 GB of GPU memory, whereas Edge Pruning requires 33 GB of GPU memory and approximately 3.5 hours for comparable performance. This substantial reduction in memory footprint enables PAHQ to run on modest-memory GPUs, making it significantly more accessible in resource-limited environments.

While Edge Pruning can achieve lower KL divergence with sufficient training steps and computational resources, PAHQ offers a compelling trade-off between performance and practical constraints. The modest increase in KL divergence (0.83 vs. 0.19-0.27 for Edge Pruning at 3000 steps) is offset by dramatic improvements in memory efficiency and runtime, making circuit discovery accessible to researchers with limited computational resources.

\section{Quantization Strategy}
\label{appendix:quantization_ablation}

This appendix provides a detailed ablation study comparing PAHQ's selective quantization strategy against full-network quantization approaches to demonstrate the effectiveness of our targeted precision preservation.

\subsection{Experimental Design}

To validate PAHQ's quantization strategy, we conducted a systematic comparison between selective quantization (our approach) and global quantization methods. The experiment follows a two-phase incremental quantization protocol:

\textbf{Phase 1: Non-critical Component Quantization}
We begin with the full-precision circuit discovered by ACDC. Following the model's reverse topological order, we iteratively select additional attention heads classified as non-critical and quantize their weights from FP32 to FP8, while preserving all critical components in FP32 precision. We record the overall task accuracy after each incremental quantization step.

\textbf{Phase 2: Critical Component Quantization}
After all non-critical attention heads have been quantized to FP8, we systematically quantize the remaining critical attention head weights from FP32 to FP8 in 10\% increments until all weights are quantized. At each step, we measure the impact on task performance to understand the sensitivity of critical components to precision reduction.
\begin{figure}[htb]
    \centering
    \includegraphics[width=\linewidth]{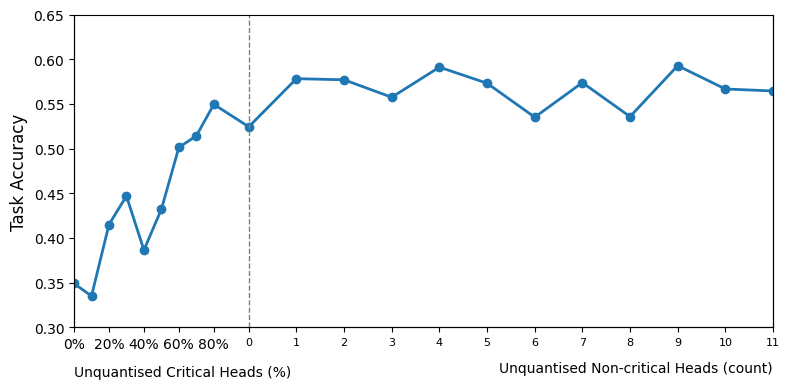}

    \caption{The results of quantization strategy comparison.}
    \label{fig:quantization}
\end{figure}
\subsection{Results and Analysis}

\begin{figure*}[htb]
    \centering
    \includegraphics[width=0.8\linewidth]{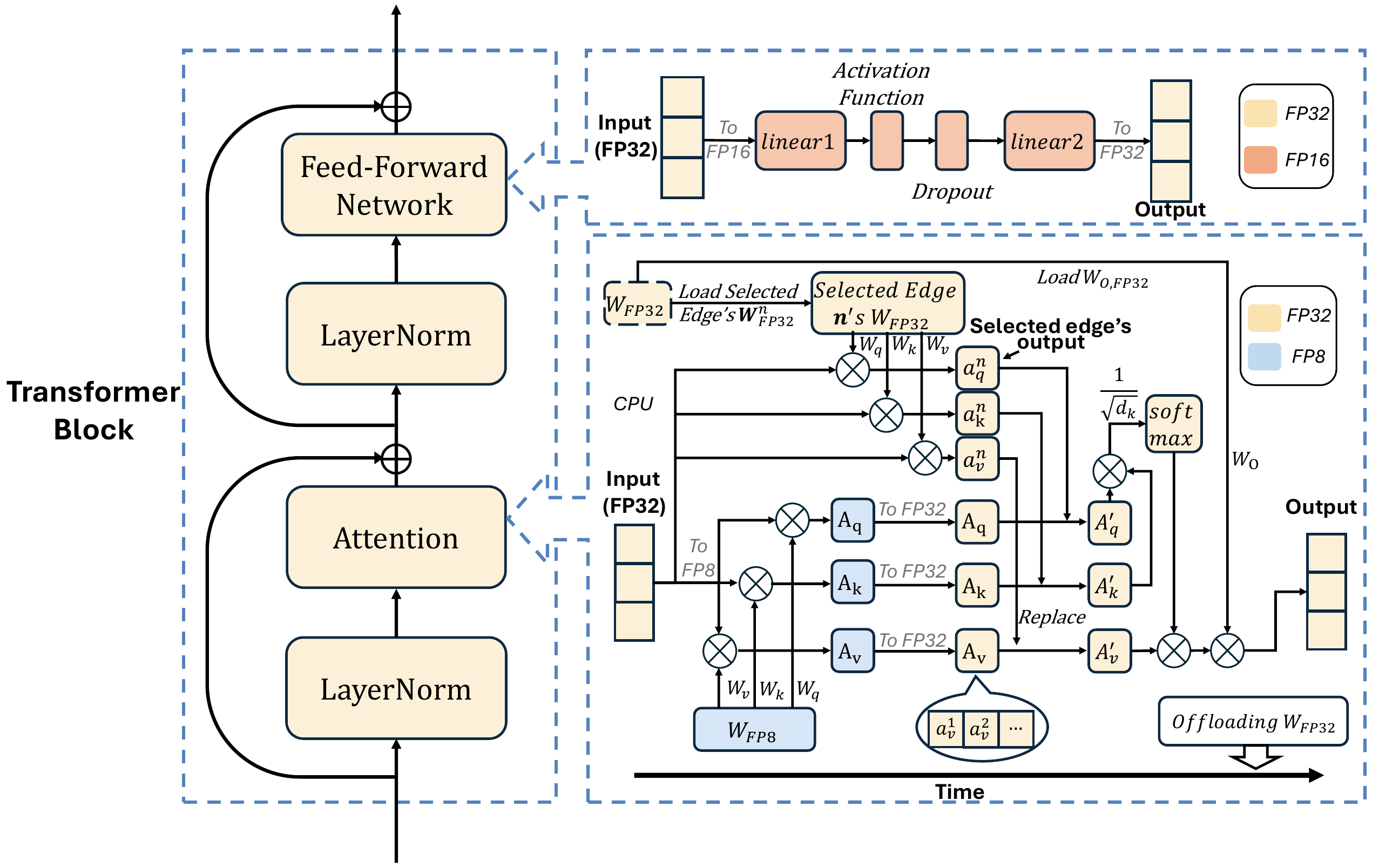}

    \caption{The full algorithm and implementation of PAHQ.}
    \label{fig:method}
\end{figure*}

Figure~\ref{fig:quantization} presents the results of our quantization strategy comparison. The experimental results confirm two key findings:

\begin{enumerate}
\item  A uniform 8-bit quantization approach, which quantizes both critical and non-critical components simultaneously, results in substantial accuracy degradation. This demonstrates that treating all network components equally during quantization is suboptimal for maintaining task performance.

\item  PAHQ's targeted preservation of FP32 precision on critical components while quantizing non-critical components to FP8 is essential for maintaining high accuracy under memory-constrained scenarios. This selective approach allows us to achieve significant memory savings without compromising the faithfulness of circuit discovery.
\end{enumerate}

These results provide empirical validation for PAHQ's core design principle: the importance of maintaining high precision for components identified as critical by circuit discovery methods while safely reducing precision elsewhere. This targeted approach enables substantial computational savings without sacrificing the reliability of mechanistic interpretability analysis.

The findings also highlight the limitations of naive quantization approaches in the context of interpretability methods, where the preservation of critical pathway faithfulness is paramount. Our selective quantization strategy offers a principled solution to this challenge, making large-scale circuit discovery more computationally feasible.

\section{Implement: Integration with ACDC}
\label{appendix:pahq_acdc}
In this section, we provide a detailed demonstration of how PAHQ readily integrates with the ACDC method. As mentioned in ~\ref{circuit discovery}, ACDC is a powerful circuit discovery technique that identifies critical connections within the model by analyzing the impact of activation replacements on downstream computations. The algorithm below describes the implementation details of the PAHQ-ACDC integration, as shown in the Figure~\ref{fig:method}.

The specific algorithm is as follows:

\begin{algorithmic}[1]
\small
\Require Model with $L$ layers and $H$ heads per layer, clean input $X$, corrupted input $\tilde{X}$, initial computational graph $G = (V, E)$, pruning threshold $\tau$, activation replacement threshold $\delta$, maximum iteration steps $T$, change rate threshold $\epsilon$, weights $W_{comp,prec}$ where $comp \in \{Q,K,V,O\}$ and $prec \in \{8,32\}$. All 8-bit weights are pre-loaded on GPU.
\Ensure Sparse computational graph $G_{sparse}$ and final output

\State Initialize scheduler $S\{\cdot\}$ to control high-precision computation $C_{high}(\cdot)$, low-precision computation $C_{low}(\cdot)$ and high-precision weight uploading $U(\cdot)$
\State Initialize computational graph $G = (V, E)$, where $V$ represents all attention heads and $E$ represents all possible edges
\State Initialize importance scores $I(e) = 0$ for all $e \in E$
\State Initialize current iteration step $t = 0$
\State Initialize previous computational graph $G_{prev} = G$

\While{$t < T$ and $|E| > 0$ and $\frac{|E_{prev}| - |E|}{|E_{prev}|} > \epsilon$}
    \State $E_{prev} \leftarrow E$
    \For{each edge $e \in E$ from source $(l_s, h_s)$ to destination $(l_d, h_d)$}
        \State // First inference: Get activations using clean input
        \State $A_{clean} \leftarrow$ Perform complete model inference with clean input $X$, record source activation $A_{clean}^{(l_s, h_s)}$
        
        \State // Second inference: Get activations using corrupted input
        \State $A_{corrupt} \leftarrow$ Perform complete model inference with corrupted input $\tilde{X}$, record source activation $A_{corrupt}^{(l_s, h_s)}$
        
        \State // Third inference: Perform inference with activation replacement
        \State Initialize dictionary for replaced activations $A_{replaced}$
        \For{each layer $l \in \{1,...,L\}$}
            \State $h^* \leftarrow 
            \begin{cases}
            h_s, & \text{if } l = l_s \\
            -1, & \text{otherwise}
            \end{cases}$
            
            \If{$h^* \neq -1$}
                \State Asynchronously launch $S\{U(W^{(l,h^*)}_{QKV,32}), U(W^{(l)}_{O,32})\}$
            \EndIf
            
            \For{$comp \in \{Q, K, V\}$}
                \State Asynchronously launch $S\{C_{low}(\tilde{X}^{(l)}, W^{(l)}_{comp,8})\}$
                \If{$h^* \neq -1$}
                    \State Synchronously wait until $U(W^{(l,h^*)}_{comp,32})$ completes
                    \State Asynchronously launch $S\{C_{high}(\tilde{X}^{(l)}, W^{(l,h^*)}_{comp,32})\}$
                \EndIf
            \EndFor

            \For{$comp \in \{Q, K, V\}$}
                \State Synchronously wait until $C_{low}(\tilde{X}^{(l)}, W^{(l)}_{comp,8})$ completes
                \If{$h^* \neq -1$}
                    \State Synchronously wait until $C_{high}(\tilde{X}^{(l)}, W^{(l,h^*)}_{comp,32})$ completes
                \EndIf
                
                \For{each head $h \in \{1,...,H\}$}
                    \State $A_{comp}^{(l,h)} = 
                    \begin{cases}
                    A_{comp,32}^{(l,h)}, & \text{if } h = h^* \text{ and } h^* \neq -1 \\
                    A_{comp,8}^{(l,h)}, & \text{otherwise}
                    \end{cases}$
                \EndFor
                \State $A_{comp} \leftarrow \text{float32}([A_{comp}^{(l,1)}; A_{comp}^{(l,2)}; ...; A_{comp}^{(l,H)}])$
            \EndFor

            \State $Output \leftarrow \text{Softmax}(A_Q \cdot A_K^T / \sqrt{d_k}) \cdot A_V$
            
            \If{$h^* \neq -1$}
                \State Synchronously wait until $U(W^{(l)}_{O,32})$ completes 
                \State $Y^{(l)} \leftarrow Output \cdot W_{O,32}$
                
                \If{$l = l_s$ and $h = h_s$}
                    \State // Replace corrupted activation with clean activation
                    \State $Y^{(l,h_s)} \leftarrow A_{clean}^{(l_s, h_s)}$
                \EndIf
            \Else
                \State $Y^{(l)} \leftarrow Output \cdot W_{O,8}$
            \EndIf
            
            \State $\tilde{X}^{(l+1)} \leftarrow$ Apply remaining operations (layer norm, MLP, etc.) to $Y^{(l)}$
            \State Store replaced activations: $A_{replaced}^{(l)} \leftarrow Y^{(l)}$
        \EndFor
        
        \State Obtain model output with replaced activations $A_{output}$
        
        \State // Calculate difference before and after replacement
        \State $\Delta A = ||A_{corrupt}^{(l_d, h_d)} - A_{replaced}^{(l_d, h_d)}||$
        
        \If{$\Delta A > \delta$}
            \State // If difference exceeds threshold, edge $e$ is considered important
            \State $I(e) \leftarrow \Delta A$
        \Else
            \State $I(e) \leftarrow 0$ // Edge is not important
        \EndIf
    \EndFor
    
    \State Sort edges based on importance scores
    \State Remove edges with importance scores below threshold: $E \leftarrow \{e \in E | I(e) > \tau\}$
    \State Update computational graph: $G \leftarrow (V, E)$
    \State $t \leftarrow t + 1$
    \State Dynamically adjust threshold $\tau$ to control pruning speed
\EndWhile

\State Perform inference using final pruned computational graph $G_{sparse} = (V, E)$:
\For{each layer $l \in \{1,...,L\}$}
    \State Determine set of heads $H_{high}^{(l)}$ requiring high-precision computation (based on edges in $G_{sparse}$)
    \For{each head $h^* \in H_{high}^{(l)}$ requiring high-precision computation}
        \State Asynchronously launch $S\{U(W^{(l,h^*)}_{QKV,32}), U(W^{(l)}_{O,32})\}$
    \EndFor
    
    \For{$comp \in \{Q, K, V\}$}
        \State Asynchronously launch $S\{C_{low}(X^{(l)}, W^{(l)}_{comp,8})\}$
        \For{each $h^* \in H_{high}^{(l)}$}
            \State Synchronously wait until $U(W^{(l,h^*)}_{comp,32})$ completes
            \State Asynchronously launch $S\{C_{high}(X^{(l)}, W^{(l,h^*)}_{comp,32})\}$
        \EndFor
    \EndFor
    
    \State Perform mixed-precision computation and merge results
    \State Proceed to next layer computation
\EndFor

\State \Return Sparse computational graph $G_{sparse}$ and final output

\end{algorithmic}
\section{Artifacts and Reproducibility}

\subsection{Software Dependencies and Licenses}

Our implementation builds upon several open-source software libraries, all of which are licensed for academic and commercial use:

\textbf{Deep Learning Frameworks:}
\begin{itemize}
    \item \textbf{PyTorch} (BSD-3-Clause License): We use PyTorch as the primary deep learning framework for model implementation and tensor operations.
    \item \textbf{Hugging Face Transformers} (Apache 2.0 License): Used for loading pre-trained language models and tokenizers. All models are accessed through the Hugging Face Model Hub\footnote{\url{https://huggingface.co/models}}.
\end{itemize}

\textbf{Mechanistic Interpretability Tools:}
\begin{itemize}
    \item \textbf{TransformerLens}~\citep{nanda2022transformerlens} (MIT License): A library specifically designed for mechanistic interpretability research, providing utilities for analyzing transformer activations and circuit discovery.
\end{itemize}

All dependencies use permissive open-source licenses that permit modification and redistribution, ensuring full reproducibility of our work.

\subsection{Models and Datasets}

\subsubsection{Pre-trained Models}

We evaluate PAHQ on three publicly available transformer models:

\begin{itemize}
    \item \textbf{GPT-2}~\citep{radford2019language}: A 124M parameter autoregressive language model trained on WebText. Available under a modified MIT license through Hugging Face (\texttt{gpt2}).
    \item \textbf{Attn-4l}~\citep{heimersheim2023circuit}: A 4-layer attention-only transformer designed for interpretability research. Publicly available under MIT license.
    \item \textbf{Redwood-2l}~\citep{Gokaslan2019OpenWeb}: A 2-layer transformer model trained on OpenWebText. Available under Apache 2.0 license.
\end{itemize}

All models were downloaded from Hugging Face Model Hub using their official model identifiers. We did not modify model weights or architectures; all experiments use the original pre-trained checkpoints.

\subsubsection{Evaluation Tasks and Datasets}

We evaluate on three standard mechanistic interpretability benchmarks:

\begin{itemize}
    \item \textbf{Indirect Object Identification (IOI)}~\citep{wang2023interpretability}: Tests the model's ability to identify indirect objects in sentences. The dataset consists of 1,000 synthetic examples with templates like "When [Name1] and [Name2] went to the store, [Name2] gave a gift to [Name1]." We converted the original token-based format to natural language and re-tokenized using each model's vocabulary.
    
    \item \textbf{Docstring}~\citep{heimersheim2023circuit}: Evaluates Python function documentation generation from code. The dataset contains 10,000 function-docstring pairs from open-source Python repositories. We use the publicly released version from the ACDC repository.
    
    \item \textbf{Greater Than}~\citep{nanda2023progress}: A numerical reasoning task requiring comparison of two numbers. The dataset includes 2,000 number pairs with varying magnitudes and digit lengths.
\end{itemize}

All datasets are either synthetically generated or derived from publicly available code repositories, containing no personally identifiable information.

\subsection{Data Privacy and Ethical Considerations}

\textbf{Pre-training Data:} The pre-trained models we use (GPT-2, Attn-4l, Redwood-2l) were trained on large-scale internet text corpora. While we did not participate in the original data collection or model training, we acknowledge:
\begin{itemize}
    \item These corpora may contain personal information that was publicly available on the internet at the time of collection
    \item The training data may include content that could be considered offensive or biased
    \item We use these models solely for analyzing their internal mechanisms and do not generate or analyze content targeting individuals or groups
\end{itemize}

\textbf{Evaluation Data:} Our evaluation datasets are either:
\begin{itemize}
    \item Synthetically generated (IOI, Greater Than) with no real personal information
    \item Derived from open-source code repositories (Docstring) with public licenses
\end{itemize}

No additional data collection or human subject involvement occurred in this research.

\subsection{Intended Use and Compatibility}

PAHQ is designed specifically for mechanistic interpretability research and maintains compatibility with the original intended uses of all utilized artifacts:

\begin{itemize}
    \item \textbf{PyTorch and Transformers}: Used within their designed scope for deep learning research and inference
    \item \textbf{TransformerLens}: Applied precisely for its intended purpose of mechanistic interpretability analysis
    \item \textbf{Pre-trained models}: Used only for analyzing internal mechanisms, not for downstream generation tasks
    \item \textbf{Circuit discovery methods}: PAHQ accelerates existing methods (ACDC) without modifying their algorithmic logic or intended research applications
\end{itemize}

Our method introduces quantization as an inference optimization technique while preserving the faithfulness and interpretability guarantees of the underlying circuit discovery algorithms. All experiments operate within the research contexts intended by the original artifact creators.

\subsection{Code and Implementation Details}

Our implementation will be made publicly available upon publication to facilitate reproduction and extension of this work. The codebase includes:

\begin{itemize}
    \item Custom CUDA kernels for mixed-precision attention computation
    \item Three-stream parallel scheduler implementation
    \item Integration with ACDC and other circuit discovery methods
    \item Evaluation scripts for all three benchmark tasks
    \item Detailed documentation and usage examples
\end{itemize}

\textbf{Hardware Requirements:} All experiments were conducted on NVIDIA H20-NVLink GPUs (96GB) with 16 vCPU AMD EPYC 9K84 processors. The method is compatible with any CUDA-capable GPU supporting FP8 operations (compute capability >=8.9).

\textbf{Software Requirements:}
\begin{itemize}
    \item Python >= 3.8
    \item PyTorch >= 2.0 with CUDA support
    \item Transformers >= 4.30
    \item TransformerLens >= 1.0
\end{itemize}

Complete installation instructions and dependency specifications will be provided in the code repository.


\end{document}